\documentclass[letterpaper]{article} 
\usepackage{aaai2026}  
\usepackage{times}  
\usepackage{helvet}  
\usepackage{courier}  
\usepackage[hyphens]{url}  
\usepackage{graphicx} 
\usepackage{natbib}  
\usepackage{caption} 
\usepackage{algorithm}
\usepackage{algorithmic}

\usepackage{multirow}
\usepackage{amsmath}
\usepackage{amsfonts}

\usepackage{newfloat}
\usepackage{listings}
\DeclareCaptionStyle{ruled}{labelfont=normalfont,labelsep=colon,strut=off} 
\floatstyle{ruled}
\newfloat{listing}{tb}{lst}{}
\floatname{listing}{Listing}
\title{Sliding-Window Merging for Compacting Patch-Redundant Layers in LLMs}

\author {
    Xuan Ding\textsuperscript{\rm 1,\rm 2}, \quad
    Rui Sun\textsuperscript{\rm 1, \rm 2}\thanks{Corresponding author.}, \quad
    Yunjian Zhang\textsuperscript{\rm 3}, \quad
    Xiu Yan\textsuperscript{\rm 4}, \quad
    Yueqi Zhou\textsuperscript{\rm 5}, \quad
    Kaihao Huang\textsuperscript{\rm 5}, \quad
    Suzhong Fu\textsuperscript{\rm 1,\rm 2}, \quad
    Angelica I Aviles-Rivero\textsuperscript{\rm 6}, \quad
    Chuanlong Xie\textsuperscript{\rm 5}, \quad
    Yao Zhu\textsuperscript{\rm 7}\footnotemark[1] 
}
\affiliations {
     \textsuperscript{\rm 1}Shenzhen Future Network of Intelligence Institute and Guangdong Provincial Key Laboratory of Future Networks of
Intelligence, The Chinese University of Hong Kong (Shenzhen), \\
     \textsuperscript{\rm 2}School of Science and Engineering, The Chinese University of Hong Kong (Shenzhen), \\
     \textsuperscript{\rm 3}University of Chinese Academy of Sciences, \\
     \textsuperscript{\rm 4}Meituan Group, \\
     \textsuperscript{\rm 5}Beijing Normal University, \\
     \textsuperscript{\rm 6}Tsinghua University, \\
     \textsuperscript{\rm 7}Zhejiang University
 
    202322011119@mail.bnu.edu.cn, ruisun@link.cuhk.edu.cn, ee\_zhuy@zju.edu.cn
}

\begin{document}

\maketitle

\begin{abstract}
Depth-wise pruning accelerates LLM inference in resource-constrained scenarios but suffers from performance degradation due to direct removal of entire Transformer layers. This paper reveals ``Patch-like'' redundancy across layers via correlation analysis of the outputs of different layers in reproducing kernel Hilbert space, demonstrating consecutive layers exhibit high functional similarity. Building on this observation, this paper proposes \textbf{S}liding-\textbf{W}indow \textbf{M}erging (SWM) - a dynamic compression method that selects consecutive layers from top to bottom using a pre-defined similarity threshold, and compacts patch-redundant layers through a parameter consolidation, thereby simplifying the model structure while maintaining its performance. Extensive experiments on LLMs with various architectures and different parameter scales show that our method outperforms existing pruning techniques in both zero-shot inference performance and retraining recovery quality after pruning. In particular, in the experiment with 35\% pruning on the Vicuna-7B model, our method achieved a 1.654\% improvement in average performance on zero-shot tasks compared to the existing method. Moreover, we further reveal the potential of combining depth pruning with width pruning to enhance the pruning effect. 
\end{abstract}

\begin{links}
    \link{Code}{https://github.com/920927/Sliding-Window-Merging}
    \link{Extended version}{https://arxiv.org/pdf/2502.19159}
\end{links}

\section{Introduction}

Large language models (LLMs) have demonstrated remarkable capabilities across a wide range of applications \citep{touvron2023llama,chowdhery2023palm,wang2025towards}, but their ever-increasing scale has also introduced significant deployment challenges. Parameter pruning is considered an effective approach to mitigate redundancy in these over-parameterized systems \citep{ma2023llm,sun2024unleashing}. Width-wise pruning removes coupled components—such as attention heads and their associated weight connections—while preserving network depth \citep{ma2023llm,an2024fluctuation,sun2024transformer}, but it struggles to address computational bottlenecks along the temporal dimension. In contrast, depth-wise pruning reduces network depth by entirely removing certain layers \citep{kim2024shortened,men2024shortgpt,song2024sleb}, significantly improving inference efficiency in resource-constrained settings. However, existing depth-wise pruning methods often lack systematic analysis of inter-layer correlations, making it difficult to fully exploit potential redundancy within the model's hierarchical structure. This may lead to structural fragmentation, disrupting the continuity of knowledge flow and ultimately resulting in notable performance degradation.

To resolve this limitation, we focus on the correlations between the layers in LLMs and pose a targeted research question: Do consecutive layers in LLMs exhibit functional redundancy beyond geometric proximity? Inspired by \citep{raghu2021vision}, we assess the correlations between the outputs of different layers of the model within a reproducing kernel Hilbert space and normalize the evaluation metric to ensure isotropic scaling invariance. Through systematic evaluation across multiple models and datasets, we reveal a consistent ``patch-redundancy" phenomenon: a high degree of similarity in the representations of certain consecutive Transformer layers in large language models, exhibiting a clear ``patch-like'' structure. This observation provides new insights for model compression, suggesting that for ``patch-redundant" layers, parameter consolidation by layer merging rather than arbitrary removal can preserve knowledge integrity while eliminating structural redundancy.

Building on the ``patch-redundancy" discovery, we propose Sliding-Window Merging (SWM) - a novel compressing method that compacts the consecutive redundant layers. This method dynamically selects the base layer and its adjacent layers with similar representations for merging, starting from the deepest layers and moving progressively towards the shallower layers, utilizing a sliding window mechanism. For the layers within the window to be merged, we calculate the parameter differences between them and the base layer, incorporating these differences into the base layer's parameters, thereby merging multiple layers into one. The sliding window mechanism selects adjacent layers of the base layer for merging by comparing the similarity between the outputs of the pruned model and the original model. When the similarity exceeds a predefined threshold, the window expands towards the shallower layers; when the similarity falls below the threshold, the layers to be merged are combined, and the window slides to update the base layer's index. Once the iterative merging is completed, a fast recovery phase is performed, utilizing limited data to post-train the pruned model.

Extensive experiments, encompassing zero-shot performance comparisons with baseline methods as well as evaluations of inference speed and throughput, demonstrate that our Sliding-Window Merging method consistently outperforms existing approaches in both model accuracy and computational efficiency. Moreover, we introduce an innovative fusion of width-wise and depth-wise pruning techniques, which further enhances the model compress performance.

The contributions of this study are summarized as:
\begin{itemize}
\item We analyze the inter-layer correlations in LLMs within a reproducing kernel Hilbert space, observing an interesting ``Patch-Like'' correlation distribution,  offering useful insights for the design of model compression strategies.
\item We propose the Sliding-Windows Merging method, which dynamically merges layers with strong representational similarity in LLMs. This method can be seamlessly applied to various LLM architectures.
\item We conduct extensive experiments across multiple LLM architectures of varying scales, demonstrating that our method outperforms existing depth-wise pruning methods in zero-shot performance, both in retraining-free scenarios and in scenarios where pruning is followed by retraining to restore quality. Specifically, when pruning the Vicuna-7B model by 35\%, our method achieved superior average performance across multiple datasets, outperforming method LLM-Pruner by 1.654\%.
\end{itemize}

\section{Related Work}

Large language models' multi-layer Transformer architecture often contains substantial redundancy, motivating research on width-wise and depth-wise pruning to reduce this redundancy and improve model efficiency.

\textbf{Width-wise pruning} reduces the network width by pruning coupled structures. For example, \citet{voita2019analyzing} and \citet{michel2019sixteen} introduced pruning and attention head sharing techniques to reduce redundant attention heads, thereby decreasing both computational complexity and parameter requirements. \citet{nova2023gradient} and \citet{santacroce2023matters} optimized the feedforward network by reducing the dimension of the FFN hidden layer, thereby reducing the memory footprint and computational complexity. More complex hybrid optimization methods have also been explored \citep{lagunas2021block,kwon2022fast,kurtic2024ziplm}. Width pruning reduces FLOPs but fails to address temporal bottlenecks in autoregressive inference, limiting latency improvements.

\textbf{Depth-wise pruning} directly removes the entire least important layer and can significantly accelerate inference. Shortened-LLM \citep{kim2024shortened} selected Taylor+ and PPL indicators as the importance measure of the Transformer layer, and deleted the unimportant Transformer layer to reduce the consumption of computing resources and improve inference speed. The layer-skipping strategy \citep{schuster2022confident,del2023skipdecode,raposo2024mixture} further reduces computational burden and boosts inference efficiency by dynamically selecting which layers to skip during execution. Additionally, \citet{song2024sleb} and \citet{tang2024rethinking} investigated depth pruning methods, which reduce model depth by eliminating redundant layers, optimizing both computational overhead and model performance while retaining essential layers. Existing depth pruning relies on heuristic importance scores, which ignore functional redundancy and cause irreversible knowledge loss through layer deletion.

\section{Method}
\subsection{Motivation}

\subsubsection{CKA vector similarity}

Center Kernel Alignment (CKA) is a metric used to compare the internal representations of neural networks. Its main advantages are its invariance to orthogonal transformations (e.g. changes in neuron arrangement) and its robustness to isotropic scaling achieved through a normalization term \citep{raghu2021vision}. These properties make CKA particularly suitable for studying the underlying relationships between different Transformer layers within large language models. Our calculation procedure for CKA is outlined as follows:

\begin{itemize}
\item Step1:  Calculate the Gram matrix of two representation matrices to measure the similarity of representations.
\begin{equation}
K=XX^T, L=YY^T, K,L \in R^{n \times n},
\label{eq1}
\end{equation}
where $X \in \mathbb{R}^{n \times p}$ and $Y \in \mathbb{R}^{n \times q}$ denote the outputs of the two Transformer layers for which CKA is to be computed, $n$ is the number of samples, and $p$ and $q$ represent the dimensionalities of $X$ and $Y$, respectively.

\item Step2: Centralize the Gram matrix to eliminate the potential impact of sample distribution deviation.
\begin{equation}
\tilde{K}=HKH, \tilde{L}=HLH,
\label{eq2}
\end{equation}
where $H=I_n-1/n 1_n 1_n^T$ is the centralization matrix, $I_n$ is the $n\times n$ identity matrix, and $1_n$ is an all-ones vector of length $n$.

\item Step3: Calculate the normalized alignment between the central Gram matrices K and L to get CKA. 
\begin{equation}
\text{CKA}(K, L) = \frac{\langle \tilde{K}, \tilde{L} \rangle_F}{\| \tilde{K} \|_F \| \tilde{L} \|_F},
\label{eq3}
\end{equation}
where $\langle \cdot, \cdot \rangle_F$ denotes the Frobenius inner product and $\| \cdot \|_F$ represents the Frobenius norm.

\end{itemize}

The final CKA value is between 0 and 1. The closer the value is to 1, the more similar the two representation matrices are.

\subsubsection{Patch-Redundancy Mechanism in LLMs}
We begin our investigation by leveraging the CKA metric to examine the internal representation structures of various models, with a particular focus on two key questions: What are the internal relationships between different Transformer layers in LLMs? And is there redundancy among these layers? 

\begin{figure}[t]
\centering
\includegraphics[width=1.0\columnwidth]{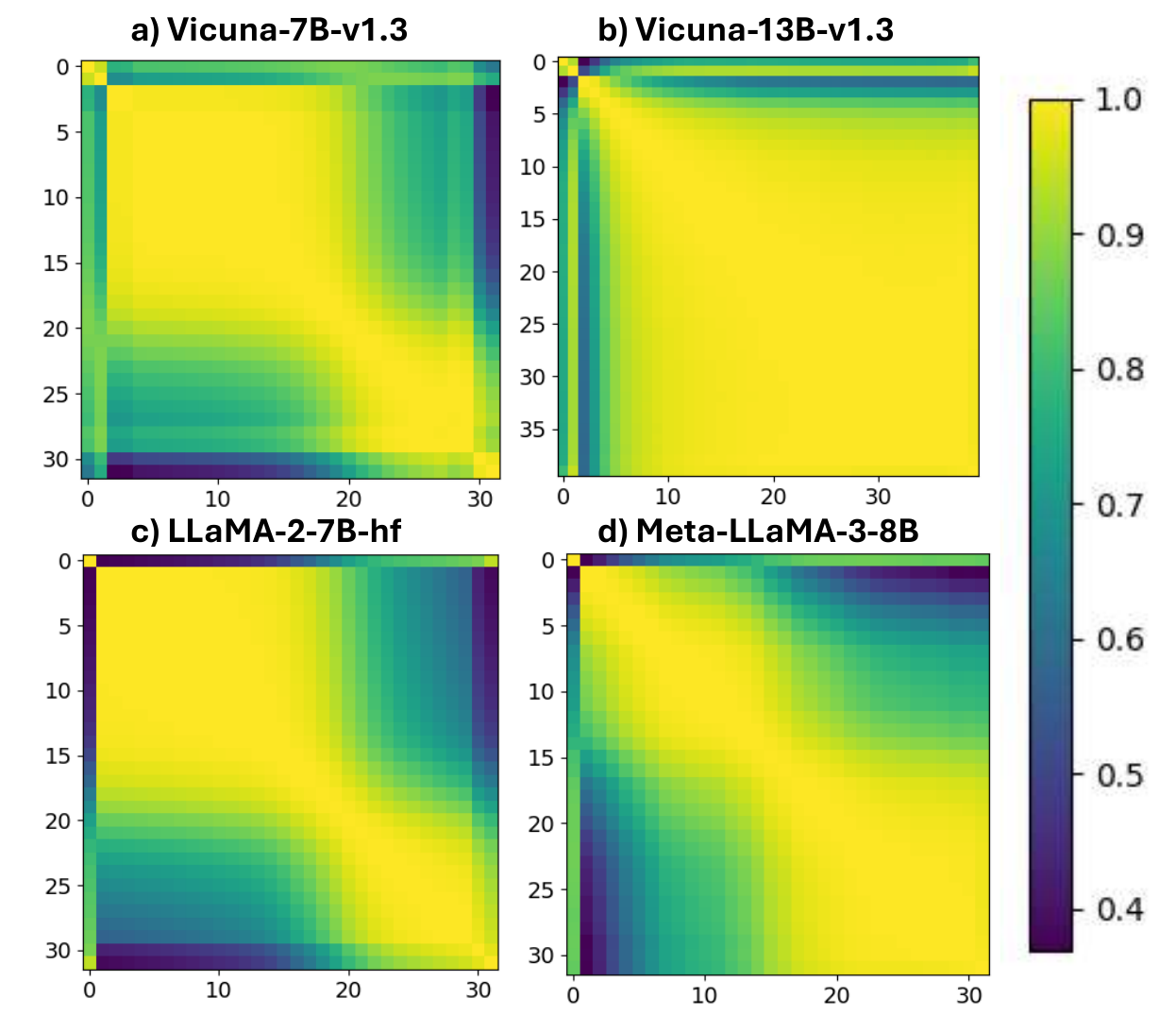}  
\caption{CKA (Center Kernel Alignment) metric between pairs of Transformer layers in LLMs.}
\label{cka}
\end{figure}

To explore these questions, we present inter-layer CKA similarity heatmaps for several LLMs, including LLaMA2-7B, Vicuna-7B-v1.3, Vicuna-13B-v1.3, and Meta-LLaMA3-8B, as shown in Fig.\ref{cka}. The results reveal a consistent geometric pattern across all evaluated LLMs: strong inter-layer correlations between adjacent intermediate layers, which appear as bright patches on the heatmaps, forming ``patch-like'' structures. This result implies that these layers have high functional redundancy and provide space for compression. 

\textbf{Theoretical Basis:} We attribute this phenomenon to two inherent properties of Transformer architectures:
\begin{enumerate}
    \item \textbf{Residual Learning Dynamics}: The residual connection $x_{l+1} = x_l + F(x_l, \theta_l)$ enables gradual feature refinement. When consecutive layers $l$ and $l+1$ satisfy $\|F(x_l, \theta_l)\| \ll \|x_l\|$, their outputs become linearly correlated, manifesting as high CKA similarity.
    \item \textbf{Optimization-Induced Redundancy}: During pretraining, layers initialized near identity transform accumulate similar gradients ($\frac{\partial \mathcal{L}}{\partial \theta_l} \approx \frac{\partial \mathcal{L}}{\partial \theta_{l+1}}$), converging to functionally equivalent solutions. 
\end{enumerate}

\subsection{Overview of SWM Design} 

Based on the aformentioned CKA analysis, we propose Sliding-Window Merging (SWM)—a novel compression method that dynamically consolidates Transformer layers from top to bottom while preserving task-critical representations through three core principles: 

\begin{itemize}

    \item \textbf{Parameter merging strategy} (Sec.3.3): The observed patch-redundancy indicates functional equivalence and provide space for compression. SWM replaces deletion with parameter consolidation (Eq.~(4)) to preserve knowledge flow. 

    \item \textbf{Adaptive similarity thresholding} (Sec.3.4): While CKA analysis reveals layer redundancy patterns, SWM employs computationally efficient cosine similarity (O(n) complexity) during runtime compression. Cosine similarity-driven thresholding criteria enable precision compression decisions, preserving CKA's isotropic invariance while accelerating computation.
    
    \item \textbf{Sliding-window merging algorithm} (Sec.3.5): Variable patch sizes across models necessitate layer-specific compression. SWM achieves this via dynamic window expansion (triggered by the comparison between cosine similarity and a predefined threshold). Moreover, low CKA correlation in boundary layers confirms their task-critical role. SWM excludes these layers from compression entirely (setting protected range [L, H] in Algorithm 1) to prevent irreversible performance damage. 

\end{itemize}

\begin{algorithm}[t]
    \caption{Iterative Layer Compression Algorithm}
    \textbf{Input}: Original model $M$ \\
    \textbf{Parameters}: Layer range $[L, H]$; Similarity threshold $T$; Few-shot calibration samples $D$ \\
    \textbf{Output}: Pruned model $M^*$
    
    \begin{algorithmic}[1] 
        \STATE $M^* \gets M$
        \STATE $h\_lay \gets H$
        \STATE $l\_lay \gets h\_lay - 1$
        \WHILE{$l\_lay \geq L$}
            \STATE $M_{tmp} \gets \text{Merge}(M^*, h\_lay, l\_lay)$
            \STATE $s \gets \text{Cal\_Sim}(M, M_{tmp}, D)$
            \IF {$s > T$}
                \STATE $l\_lay \gets l\_lay - 1$
            \ELSE
                \STATE $M^* \gets M_{tmp}$
                \STATE $h\_lay \gets l\_lay$
            \ENDIF
        \ENDWHILE
        \STATE \textbf{return} $M^*$
    \end{algorithmic}
\label{algorithm1}
\end{algorithm}

\begin{figure*}[htbp]
\centering
\includegraphics[width=1.0\textwidth]{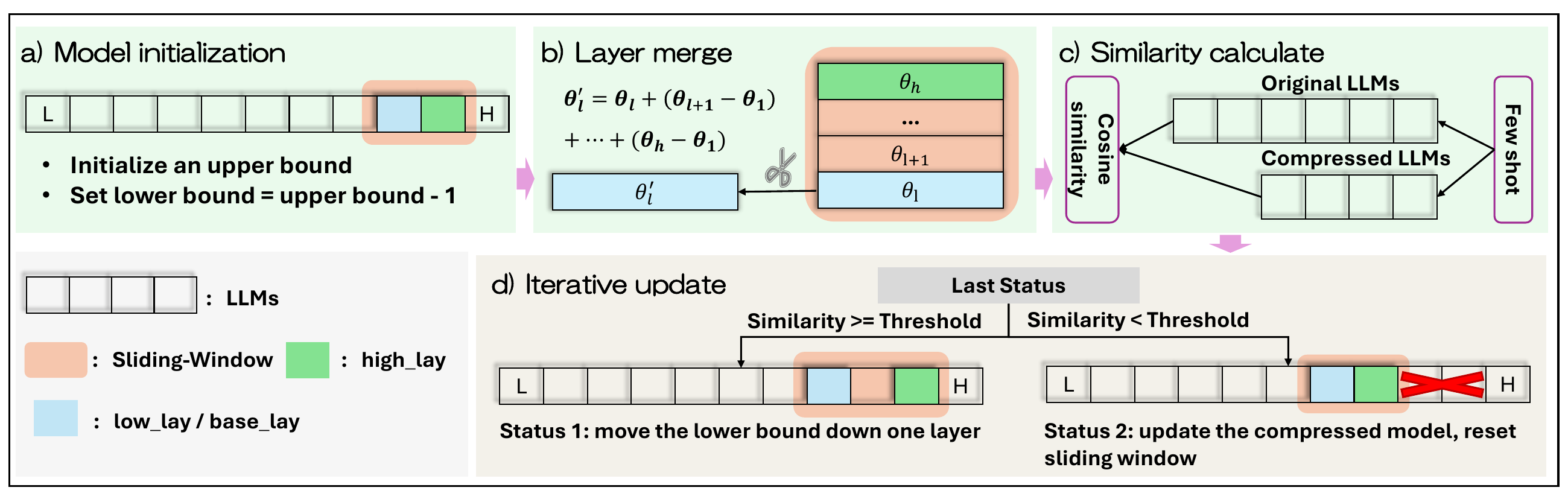}
\caption{The framework of our sliding-windows merging (SWM) method. (a) Model initialization establishes upper / lower bounds for the sliding window. (b) Layer merging via parameter consolidation: $\theta^*_l$ denotes merged parameters combining adjacent layers within the window. (c) Similarity validation computes cosine similarity between original and compressed models' outputs using few-shot evaluation. (d) Adaptive window adjustment: the window slides downward if similarity meets thresholds (Status 1), otherwise the compressed model updates and window resets (Status 2). Color coding: gray blocks (original LLM layers), red-orange window (active merging region), green / blue markers (current upper / lower bounds).}
\label{framework}
\end{figure*}

\subsection{Parameter merging strategy}

We compact layers based on inter-layer differences (as shown in Fig.\ref{framework}(b)), which progressively integrates redundant information while preserving core functionality. Specifically, given layers $\{L_i, L_{i+1}, ...,L_j\}$ with parameters $\{\theta_i, \theta_{i+1}, ...,\theta_j\}$ within the sliding window, our Merge function computes the merged parameters $\theta^*$ as:
\begin{align}
\theta_i^* &= \theta_i + (\theta_{i+1} - \theta_i) + \cdots + (\theta_{j} - \theta_i) \nonumber \\
          &= \theta_i + \sum_{k=1}^{j-i}(\theta_{i+k}-\theta_i),
\end{align}

This formulation offers two key advantages: (1) the base layer $\theta_i$ maintains fundamental model capabilities, while (2) the difference terms ($\theta_{i+k}-\theta_i$) incorporate complementary features from other layers. The strategy effectively balances knowledge preservation and redundancy elimination across varying compression requirements.

\subsection{Adaptive similarity thresholding}
The dynamic window updating process relies on comparing cosine similarity between original ($M$) and compressed ($M_{tmp}$) model outputs against predefined thresholds iteratively (Fig.\ref{framework} (c)). Through systematic ablation (Section 5.6), we observe that lower thresholds enable more aggressive compression but degrade performance, while higher thresholds better preserve model behavior at the cost of reduced compression efficiency. We optimize thresholds via validation-set grid search under pruning constraints, selecting the highest feasible value maximizing zero-shot capability retention to ensure better model performance.

\subsection{Sliding window merging algorithm}

As shown in Algorithm \ref{algorithm1}, our method performs iterative layer compression through three steps: (1) model initialization, (2) dynamic window updating, and (3) termination condition evaluation.

\textbf{Model initialization}. We initialize the target compressed model $M^*$ as a copy of the original model $M$ while simultaneously configuring the sliding window (upper=$H$, lower=$H-1$) within a predefined compression range $[L, H]$, where $H$ and $L$ denote the highest and lowest layers to be compressed, respectively (see Fig.\ref{framework} (a)). To preserve critical functionality, we exclude the top layers from compression based on CKA analysis, which reveals their distinct role due to lower inter-layer correlation.

\textbf{Dynamic window updating.} Our method iteratively merge layers within current sliding window to create $M_{tmp}$ and compute its cosine similarity with the original model $M$ using last hidden states from few-shot calibration data. When the similarity exceeds threshold $T$ (indicating minimal performance impact), we expand the merging range by moving the lower bound down one layer. Conversely, when the similarity falls below threshold $T$ (suggesting significant performance impact), we stop window expanding, update the compressed model $M^ * = M_{tmp}$, and reset the upper bound to the current lower bound before proceeding to the next merging round (see Fig.\ref{framework} (d)). This adaptive process systematically balances compression efficiency with performance preservation through representation-aware thresholding, terminating when further merging would violate the similarity constraint to maintain model integrity.

\textbf{Termination condition evaluation.} The dynamic window updating process continues until the lowest level $L$ is processed. Ultimately, the pruned model $M^*$ output by the algorithm reduces redundant computing and storage requirements by retaining the merged representation of key layers.

\subsection{Performance Recovery by Low-rank Approximation}

We use the low-rank approximation technique, LoRA \citep{hulora}, to fine-tune the pruned model and recover its performance. This is a common practice in many pruning methods \citep{ma2023llm,kim2024shortened}, and we provide a brief introduction to ensure the self-contained aspect of our work in Appendix A.3.

\begin{table*}[t]
\centering
\renewcommand{\arraystretch}{1.15}
\resizebox{2.0\columnwidth}{!}{%
\begin{tabular}{ccc|cccccccc}
\hline \hline
\multicolumn{1}{c|}{Pruned Rates} & \multicolumn{2}{c|}{Model} & BoolQ & PIQA & HellaSwag & WinoGrande & ARC-easy & ARC-challenge & OpenbookQA & AVE \\ \hline \hline
\multicolumn{1}{c|}{0\%} & \multicolumn{2}{c|}{LLaMA2-7B} & 77.706 & 78.074 & 76.021 & 69.140 & 76.305 & 46.331 & 44.200 & 66.825 \\ \hline
\multicolumn{1}{c|}{\multirow{6}{*}{20\%}} & \multicolumn{1}{c|}{\multirow{3}{*}{width}} & Wanda-sp & 62.600 & 76.440 & 70.660 & 63.770 & 69.610 & 42.150 & 40.000 & 60.747 \\
\multicolumn{1}{c|}{} & \multicolumn{1}{c|}{} & FLAP & 72.050 & 73.390 & 64.690 & 64.720 & 62.250 & 32.510 & 36.800 & 58.059 \\
\multicolumn{1}{c|}{} & \multicolumn{1}{c|}{} & LLM-Pruner & 63.731 & 77.476 & 67.128 & 61.878 & 65.783 & 38.481 & 40.400 & 59.268 \\ \cline{2-11} 
\multicolumn{1}{c|}{} & \multicolumn{1}{c|}{\multirow{3}{*}{depth}} & SLEB & 62.875 & 73.939 & 63.951 & 59.747 & 63.468 & 35.154 & 38.000 & 56.733 \\
\multicolumn{1}{c|}{} & \multicolumn{1}{c|}{} & Shortened-LLM & 61.560 & 76.061 & 67.994 & 58.800 & 68.813 & 37.884 & 38.000 & 58.445 \\
\multicolumn{1}{c|}{} & \multicolumn{1}{c|}{} & Ours & 69.450 & 73.667 & 70.484 & 67.088 & 69.108 & 41.212 & 42.600 & \textbf{61.944} \\ \hline
\multicolumn{1}{c|}{\multirow{6}{*}{35\%}} & \multicolumn{1}{c|}{\multirow{3}{*}{width}} & Wanda-sp & 59.790 & 68.820 & 53.140 & 54.060 & 52.270 & 31.570 & 32.800 & 50.350 \\
\multicolumn{1}{c|}{} & \multicolumn{1}{c|}{} & FLAP & 66.970 & 67.850 & 52.100 & 61.480 & 49.490 & 28.070 & 32.400 & 51.194 \\
\multicolumn{1}{c|}{} & \multicolumn{1}{c|}{} & LLM-Pruner & 45.260 & 74.760 & 60.290 & 59.350 & 57.280 & 32.420 & 37.200 & \textbf{52.366} \\ \cline{2-11} 
\multicolumn{1}{c|}{} & \multicolumn{1}{c|}{\multirow{3}{*}{depth}} & SLEB & 54.801 & 67.410 & 46.545 & 53.197 & 48.527 & 29.010 & 33.000 & 47.499 \\
\multicolumn{1}{c|}{} & \multicolumn{1}{c|}{} & Shortened-LLM & 62.171 & 71.926 & 54.800 & 51.776 & 59.259 & 30.887 & 35.400 & 52.317 \\
\multicolumn{1}{c|}{} & \multicolumn{1}{c|}{} & Ours & 63.119 & 65.452 & 56.503 & 58.879 & 52.020 & 31.911 & 35.200 & 51.869 \\ \hline \hline
\multicolumn{1}{c|}{0\%} & \multicolumn{2}{c|}{Vicuna-7B-v1.3} & 78.104 & 77.312 & 73.939 & 69.376 & 74.327 & 44.454 & 43.800 & 65.902 \\ \hline
\multicolumn{1}{c|}{\multirow{6}{*}{20\%}} & \multicolumn{1}{c|}{\multirow{3}{*}{width}} & Wanda-sp & 63.270 & 73.780 & 68.620 & 63.930 & 67.210 & 38.820 & 37.200 & 58.976 \\
\multicolumn{1}{c|}{} & \multicolumn{1}{c|}{} & FLAP & 73.520 & 74.810 & 68.760 & 66.460 & 69.110 & 38.990 & 40.000 & 61.664 \\
\multicolumn{1}{c|}{} & \multicolumn{1}{c|}{} & LLM-Pruner & 67.645 & 76.115 & 66.660 & 63.931 & 65.446 & 36.604 & 40.400 & 59.543 \\ \cline{2-11} 
\multicolumn{1}{c|}{} & \multicolumn{1}{c|}{\multirow{3}{*}{depth}} & SLEB & 46.758 & 51.850 & 26.170 & 51.223 & 25.505 & 28.840 & 24.800 & 36.449 \\
\multicolumn{1}{c|}{} & \multicolumn{1}{c|}{} & Shortened-LLM & 72.355 & 74.701 & 67.576 & 64.562 & 70.034 & 38.225 & 38.600 & 60.865 \\
\multicolumn{1}{c|}{} & \multicolumn{1}{c|}{} & Ours & 77.431 & 74.755 & 69.040 & 68.745 & 69.318 & 38.908 & 40.200 & \textbf{62.628} \\ \hline
\multicolumn{1}{c|}{\multirow{6}{*}{35\%}} & \multicolumn{1}{c|}{\multirow{3}{*}{width}} & Wanda-sp & 50.760 & 60.450 & 43.060 & 55.960 & 43.520 & 26.190 & 28.000 & 43.991 \\
\multicolumn{1}{c|}{} & \multicolumn{1}{c|}{} & FLAP & 57.860 & 69.640 & 59.120 & 63.300 & 57.830 & 35.670 & 36.000 & 54.203 \\
\multicolumn{1}{c|}{} & \multicolumn{1}{c|}{} & LLM-Pruner & 63.976 & 73.069 & 59.560 & 58.564 & 56.524 & 32.679 & 37.800 & 54.596 \\ \cline{2-11} 
\multicolumn{1}{c|}{} & \multicolumn{1}{c|}{\multirow{3}{*}{depth}} & SLEB & 37.829 & 53.264 & 25.921 & 49.961 & 25.926 & 29.096 & 25.800 & 35.399 \\
\multicolumn{1}{c|}{} & \multicolumn{1}{c|}{} & Shortened-LLM & 64.281 & 70.783 & 56.722 & 57.380 & 59.596 & 31.485 & 34.000 & 53.464 \\
\multicolumn{1}{c|}{} & \multicolumn{1}{c|}{} & Ours & 69.235 & 70.294 & 60.705 & 62.273 & 60.227 & 33.618 & 37.400 & \textbf{56.250} \\ \hline \hline
\multicolumn{1}{c|}{0\%} & \multicolumn{2}{c|}{LLaMA3-8B} & 81.101 & 79.489 & 79.167 & 73.402 & 80.093 & 53.242 & 44.800 & 70.185 \\ \hline
\multicolumn{1}{c|}{\multirow{5}{*}{20\%}} & \multicolumn{1}{c|}{\multirow{2}{*}{width}} & FLAP & 37.830 & 52.180 & 26.360 & 49.960 & 26.810 & 24.830 & 26.000 & 34.853 \\
\multicolumn{1}{c|}{} & \multicolumn{1}{c|}{} & LLM-Pruner & 74.037 & 79.489 & 74.268 & 70.324 & 74.285 & 46.587 & 42.600 & 65.941 \\ \cline{2-11} 
\multicolumn{1}{c|}{} & \multicolumn{1}{c|}{\multirow{3}{*}{depth}} & SLEB & 62.171 & 73.286 & 64.748 & 63.062 & 64.562 & 37.713 & 37.000 & 57.506 \\
\multicolumn{1}{c|}{} & \multicolumn{1}{c|}{} & Shortened-LLM & 66.208 & 78.074 & 72.695 & 62.747 & 75.295 & 44.795 & 43.400 & 63.316 \\
\multicolumn{1}{c|}{} & \multicolumn{1}{c|}{} & Ours & 76.789 & 77.639 & 73.770 & 71.744 & 76.599 & 50.939 & 41.200 & \textbf{66.954} \\ \hline
\multicolumn{1}{c|}{\multirow{5}{*}{35\%}} & \multicolumn{1}{c|}{\multirow{2}{*}{width}} & FLAP & 37.830 & 52.340 & 26.050 & 47.990 & 24.790 & 24.830 & 25.200 & 34.147 \\
\multicolumn{1}{c|}{} & \multicolumn{1}{c|}{} & LLM-Pruner & 64.465 & 74.048 & 61.800 & 59.353 & 64.646 & 34.386 & 37.200 & 56.557 \\ \cline{2-11} 
\multicolumn{1}{c|}{} & \multicolumn{1}{c|}{\multirow{3}{*}{depth}} & SLEB & 59.755 & 64.635 & 45.061 & 51.539 & 47.306 & 27.133 & 27.600 & 46.147 \\
\multicolumn{1}{c|}{} & \multicolumn{1}{c|}{} & Shortened-LLM & 63.180 & 72.851 & 62.985 & 58.090 & 66.877 & 37.116 & 37.000 & 56.871 \\
\multicolumn{1}{c|}{} & \multicolumn{1}{c|}{} & Ours & 67.554 & 73.830 & 61.472 & 62.747 & 64.352 & 36.007 & 37.600 & \textbf{57.652} \\ \hline \hline
\end{tabular}
}
\caption{Zero-shot performance comparison of pruning methods at 20\% and 35\% pruning rates. We compare our method with width pruning (Wanda-sp, FLAP, LLM-Pruner) and depth pruning (SLEB, Shortened-LLM) methods on LLaMA2-7B, Vicuna-7B, and LLaMA3-8B. Note: Wanda-sp does not produce results for LLaMA3-8B due to incompatibility.}
\label{table1}
\end{table*}

\section{Experiments}

\subsection{Experimental setup}
\textbf{Foundation LLMs.} We conducte experiments on existing popular open-source language models, including LLaMA2-\{7B, 13B\} \citep{touvron2023llama}, LLaMA3-\{8B\} and Vicuna-\{7B, 13B\}-v1.3 \citep{chiang2023vicuna}.

\textbf{Baselines.} We benchmark SWM against several width pruning (LLM-Pruner \citep{ma2023llm}, FLAP \& Wanda-sp \citep{an2024fluctuation}) and depth pruning (SLEB \citep{song2024sleb}, Shortened-LLM \citep{kim2024shortened}) methods. Following the experimental setup of Shortened-LLM, we assess all methods under two target pruning levels: 20\% and 35\%. If the product of the total number of Transformer layers and target sparsity is not an integer, we round up to determine the number of layers to remove.

\textbf{Benchmarks.} Following \citet{touvron2023llama}, we measure model performance on seven commonsense reasoning datasets (i.e., BoolQ \citep{clark2019boolq}, PIQA \citep{bisk2020piqa}, HellaSwag \citep{zellers2019hellaswag}, WinoGrande \citep{sakaguchi2021winogrande}, ARCeasy \citep{clark2018think}, ARC-challenge \citep{clark2018think}, and OpenbookQA \citep{mihaylov2018can}) using the lm-evaluation-harness package \citep{eval-harness}. 

\textbf{Implementation Details.} Our implementation builds on PyTorch \citep{paszke2019pytorch} and HuggingFace Transformers \citep{wolf2020transformers}. Following \citet{ma2023llm}, we randomly select 10 samples from BookCorpus \citep{zhu2015aligning} for similarity computation during pruning. We also use this calibration dataset for baselines to ensure a fair comparison. In LoRA retraining, we use 50K samples of refined Alpaca \citep{taori2023stanford} for instruction tuning. All experiments executed on NVIDIA A100 GPUs (80GB memory), with pruning efficiently integrated into forward passes (about 2 minutes for 13B models) and memory footprint comparable to standard inference.

\subsection{Zero-shot Tasks}

Tab.\ref{table1} demonstrates SWM's consistent superiority over both width and depth pruning baselines across LLaMA2-7B, Vicuna-7B-v1.3, and LLaMA3-8B models. Specifically, under the 20\% pruning rate of LLaMA2-7B model, our method achieves a 2.676\% higher accuracy than the best-performing LLM-Pruner method; under the 35\% pruning rate of Vicuna-7B model, the average accuracy of our method is 1.654\% higher than that of existing methods. Moreover, when handling advanced architectures like LLaMA3-8B, SWM maintains robust performance (66.954 vs original 70.185) while width pruning method FLAP suffers catastrophic failure (34.147 at 35\% pruning). These results show that our method can effectively reduce model size and complexity while more fully maintaining performance, attributable to its knowledge-preserving consolidation mechanism.

To verify the broad applicability of our method, we also provide relevant experimental results on larger models (such as LLaMA2-13B and Vicuna-13B-v1.3) in Appendix B.4.

\subsection{Latency and Throughput}

\begin{table}[t]
\renewcommand{\arraystretch}{1.55}
\fontsize{20}{15}\selectfont
\resizebox{1.0\columnwidth}{!}{%
\begin{tabular}{ccl|clclclcl}
\hline \hline
\multicolumn{1}{c|}{Pruned Rates} & \multicolumn{2}{c|}{Model} & \multicolumn{2}{c}{Latency} & \multicolumn{2}{c}{Throughout} & \multicolumn{2}{c}{GPU\_Mem} & \multicolumn{2}{c}{nparam} \\ \hline
\multicolumn{1}{c|}{0\%} & \multicolumn{2}{c|}{LLaMA2-7B} & \multicolumn{2}{c}{2.729} & \multicolumn{2}{c}{46.905} & \multicolumn{2}{c}{13020.25} & \multicolumn{2}{c}{6.7B} \\ \hline
\multicolumn{1}{c|}{\multirow{6}{*}{20\%}} & \multicolumn{1}{c|}{\multirow{3}{*}{width}} & Wanda-sp & \multicolumn{2}{c}{4.628} & \multicolumn{2}{c}{27.663} & \multicolumn{2}{c}{10676} & \multicolumn{2}{c}{5.5B} \\
\multicolumn{1}{c|}{} & \multicolumn{1}{c|}{} & FLAP & \multicolumn{2}{c}{4.045} & \multicolumn{2}{c}{31.656} & \multicolumn{2}{c}{10707.25} & \multicolumn{2}{c}{5.4B} \\
\multicolumn{1}{c|}{} & \multicolumn{1}{c|}{} & LLM-Pruner & \multicolumn{2}{c}{5.655} & \multicolumn{2}{c}{22.635} & \multicolumn{2}{c}{10951.5} & \multicolumn{2}{c}{5.5B} \\ \cline{2-11} 
\multicolumn{1}{c|}{} & \multicolumn{1}{c|}{\multirow{3}{*}{depth}} & SLEB & \multicolumn{2}{c}{2.529} & \multicolumn{2}{c}{50.622} & \multicolumn{2}{c}{10682.45} & \multicolumn{2}{c}{5.5B} \\
\multicolumn{1}{c|}{} & \multicolumn{1}{c|}{} & Shortened-LLM & \multicolumn{2}{c}{2.585} & \multicolumn{2}{c}{49.542} & \multicolumn{2}{c}{10682.45} & \multicolumn{2}{c}{5.5B} \\
\multicolumn{1}{c|}{} & \multicolumn{1}{c|}{} & Ours & \multicolumn{2}{c}{2.339} & \multicolumn{2}{c}{54.758} & \multicolumn{2}{c}{10682.45} & \multicolumn{2}{c}{5.5B} \\ \hline
\multicolumn{1}{c|}{\multirow{6}{*}{35\%}} & \multicolumn{1}{c|}{\multirow{3}{*}{width}} & Wanda-sp & \multicolumn{2}{c}{4.619} & \multicolumn{2}{c}{27.726} & \multicolumn{2}{c}{8901} & \multicolumn{2}{c}{4.5B} \\
\multicolumn{1}{c|}{} & \multicolumn{1}{c|}{} & FLAP & \multicolumn{2}{c}{4.127} & \multicolumn{2}{c}{31.051} & \multicolumn{2}{c}{8855.95} & \multicolumn{2}{c}{4.5B} \\
\multicolumn{1}{c|}{} & \multicolumn{1}{c|}{} & LLM-Pruner & \multicolumn{2}{c}{5.630} & \multicolumn{2}{c}{22.736} & \multicolumn{2}{c}{9043.9} & \multicolumn{2}{c}{4.5B} \\ \cline{2-11} 
\multicolumn{1}{c|}{} & \multicolumn{1}{c|}{\multirow{3}{*}{depth}} & SLEB & \multicolumn{2}{c}{1.938} & \multicolumn{2}{c}{66.048} & \multicolumn{2}{c}{8725.9} & \multicolumn{2}{c}{4.5B} \\
\multicolumn{1}{c|}{} & \multicolumn{1}{c|}{} & Shortened-LLM & \multicolumn{2}{c}{2.084} & \multicolumn{2}{c}{61.433} & \multicolumn{2}{c}{8725.85} & \multicolumn{2}{c}{4.5B} \\
\multicolumn{1}{c|}{} & \multicolumn{1}{c|}{} & Ours & \multicolumn{2}{c}{1.889} & \multicolumn{2}{c}{67.770} & \multicolumn{2}{c}{8725.9} & \multicolumn{2}{c}{4.5B} \\ \hline \hline
\end{tabular}
}
\caption{Inference efficiency comparison of pruning methods. (Measured with 12 input tokens, 128 output tokens and a batch size of 1.)}
\label{inference_efficiency}
\end{table}

We follow \citet{sheng2023flexgenhighthroughputgenerativeinference} to evaluate the LLM inference speedup achieved by our pruning methods. Given a batch size M and an output sequence length L, the latency T represents the time required to handle the given prompts and produce $ML$ output tokens. The throughput is computed as $ML/T$ . We report the average results from 20 runs after the initial 10 warm-up batches. Tab.\ref{inference_efficiency} present throughput and latency results for LLaMA2-7B.

Experimental results show that depth pruning generally outperforms width pruning in reasoning efficiency. Specifically, at pruning ratio of 20\% and 35\%, depth pruning (SLEB, Shortened-LLM and Ours) outperform width pruning methods (Wanda-sp, FLAP, and LLM-Pruner) in both latency and throughout. This suggests that reducing model depth can more effectively enhances inference speed and throughout. Additionally, depth pruning methods maintain relatively stable GPU memory usage while ensuring efficient inference. Therefore, from the perspective of inference efficiency, depth pruning is a more effective pruning strategy.

\begin{figure}[t]
\centering
\includegraphics[width=1.0\columnwidth]{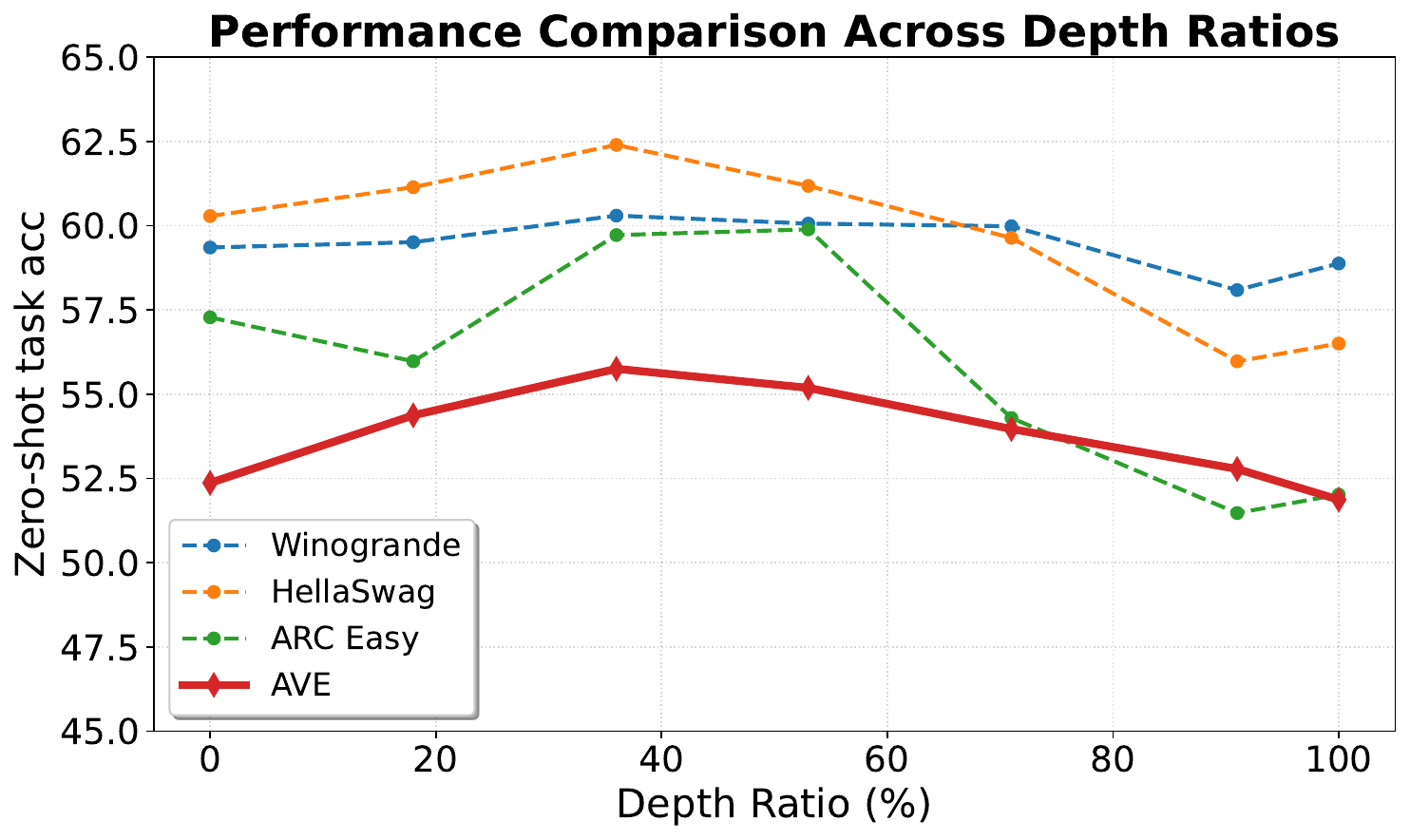}  
\caption{Performance of the integrated method on LLaMA2-7B. The horizontal axis shows the depth pruning ratio, and the vertical axis indicates zero-shot task performance. Dotted lines denote individual task metrics (Winogrande, HellaSwag, ARC-easy), while the solid line shows the average across all seven tasks.}
\label{merge}
\end{figure}

\subsection{Integration of width pruning and depth pruning}

We observed in Tab.\ref{table1} that in the 35\% pruning scenario for LLaMA2-7B, the width pruning (LLM-Pruner) slightly outperforms our depth-wise method. This may be due to the fact that under specific pruning ratios and model structures, width pruning can also exhibit advantages, and depth pruning does not necessarily always outperform it. This naturally raises the question: Can the intergration of width pruning and depth pruning leverage the strengths of both methods to further improve pruning results?

Specifically, using LLaMA2-7B model as an example, we propose a combined compression paradigm integrating SWM's layer consolidation with LLM-Pruner's structured sparsity. In the depth stage, we apply SWM to compress redundant layers of pruning attributed to depth-wise pruning (ranging from 0\% to 100\%); In the width stage, we employ LLM-Pruner on the model obtained in the depth stage to remove non-essential coupled structures, achieving total 35\% sparsity. For the sake of pruning convenience, we selected the following depth pruning rates: 0\% (LLM-Pruner), 18\%, 36\%, 53\%, 71\%, 91\%, and 100\% (SWM). 

As shown in Fig.\ref{merge}, the combined depth and width pruning strategy achieves better performance at the same pruning rate compared to using either method alone. Specifically, models with depth pruning rates of 18\%, 36\%, 53\%, 71\% and 91\% consistently surpass those with 0\% and 100\% depth pruning. Notably, models with depth pruning rates of 36\% and 53\% rank first and second in performance, respectively. This shows that the integrated methods leverage the advantages of both depth pruning and width pruning methods, achieving better model performance than when using depth pruning or width pruning alone, while mitigating the throughput and inference speed issues associated with width pruning methods. We also performed the same experiments on Vicuna-13B-v1.3 model, with results in Appendix B.2.

\begin{figure}[t]
\centering
\includegraphics[width=1.0\columnwidth]{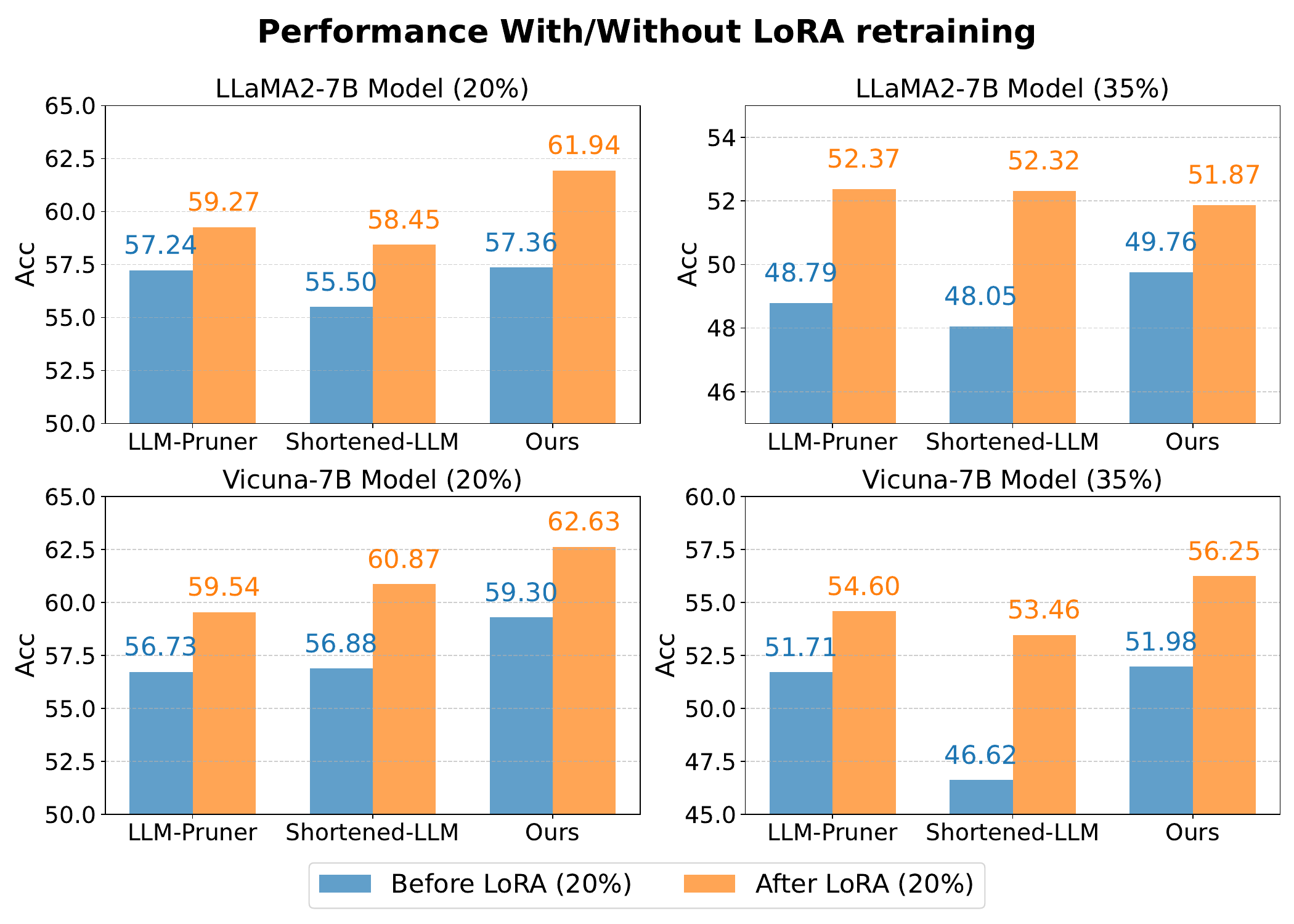}  
\caption{Performance with/without LoRA retraining. The blue column shows performance before LoRA fine-tuning, and the orange column after.}
\label{lora}
\end{figure}

\subsection{The impact of LoRA retraining}

We evaluate the impact of LoRA retraining on three baselines requiring fine-tuning (LLM-Pruner, Shortened-LLM, and ours) across LLaMA2-7B and Vicuna-7B at 20\% and 35\%  pruning rates. The results in Fig.\ref{lora} show that before LoRA fine-tuning, our method significantly outperformed LLM-Pruner and Shortened-LLM, achieving the best performance at both 20\% and 35\% pruning ratios. After further employing LoRA fine-tuning, our method's performance significantly improves and still maintains the lead in most cases. Specifically, under the 20\% pruning ratio of LLaMA2-7B model, our method improves from 57.360\% to 61.944\%, a 4.584\% increase, far exceeding LLM-Pruner (by 2.031\%) and Shortened-LLM (by 2.946\%).

\subsection{The impact of Layer merging methods}
We evaluate three layer merging methods — direct layer deletion (``Delete''), replacing multiple layers' parameters with their average (``Average''), and our method (``Ours'') — on LLaMA2-7B model, analyzing their impact on pruned model performance using the HellaSwag dataset (without LoRA fine-tuning).

\begin{figure}[t]
\centering
\includegraphics[width=1.0\columnwidth]{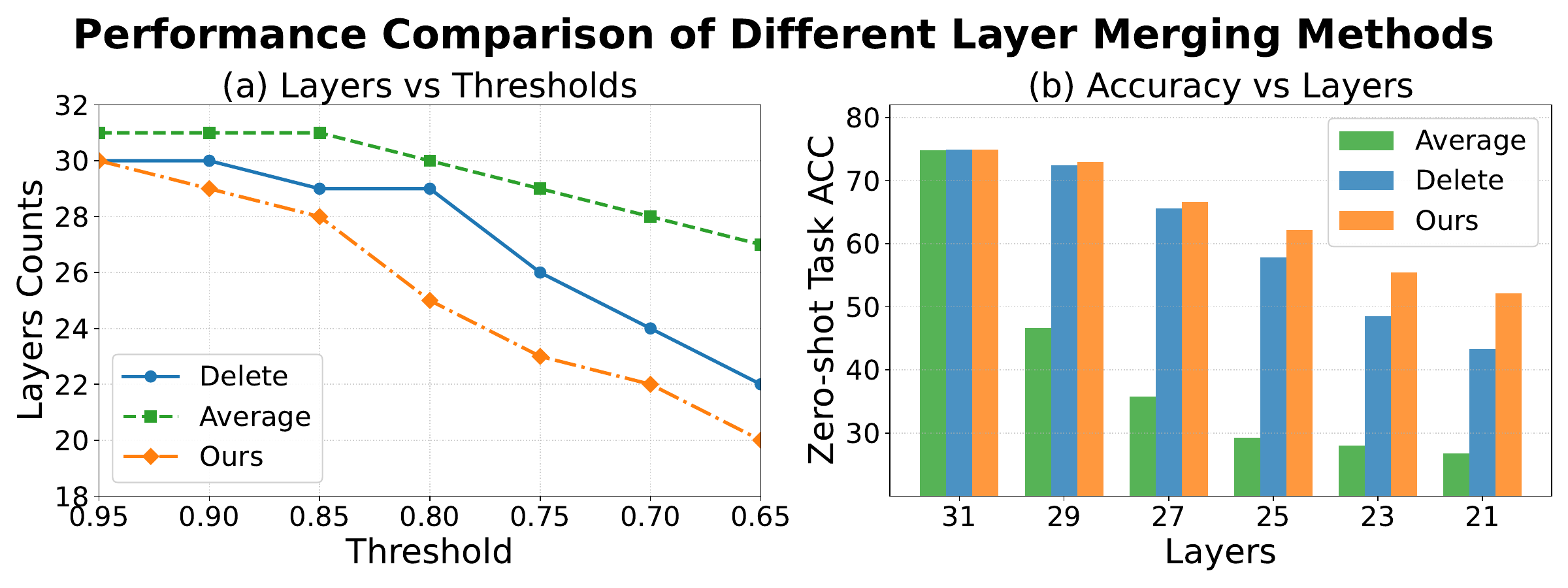}  
\caption{The impact of layer merging methods on LLaMA2-7B model. (a) Layer counts of the pruned model under different similarity thresholds through three layer merging methods: Delete (blue circles), Average (green squares), Ours (orange diamonds). (b) Zero-shot performance of the pruned model on the HellaSwag dataset through three layer merging methods: Delete (blue bars), Average (green bars), Ours (orange bars).}
\label{params}
\end{figure}

We first analyzed the layer count after different merging methods at varying similarity thresholds in Fig.\ref{params}(a). As the threshold decreases, the number of layers reduces, indicating fewer redundant parameters. At the same threshold, our layer merging method achieves higher compression. For example, at a threshold of 0.75, our method retains only 23 layers, significantly fewer than ``Delete'' method (26 layers) and ``Average'' method (29 layers), resulting in a more efficient model compression.

In Fig.\ref{params}(b), we compared the zero-shot accuracy at various compression levels after different merging methods. The results show that, while ``Average'' degrades significantly under aggressive pruning, our method consistently outperforms both baselines, with the performance gap gradually expanding as the number of compression layers increases.

\subsection{The impact of Calibration data}
Tab.\ref{calibration} investigates the impact of calibration data selection on LLaMA2-7B model. The results confirm our method's robustness to calibration data selection. Across diverse corpora and sample sizes, performance variations on tasks are minimal. Specifically, HellaSwag accuracy fluctuates within 1.6\% absolute (62.2–63.7\%), while OpenbookQA shows less than 1.0\% deviation (36.2–37.2\%). Crucially, no dataset demonstrates consistent superiority, and larger sample sizes yield diminishing returns. The observed stability suggests that choice of calibration data has minimal practical impact on our method's effectiveness. This relative invariance highlights the operational robustness of our method for real-world compression scenarios.

\begin{table}[t]
\centering
\resizebox{1.0\columnwidth}{!}{%
\begin{tabular}{c|cc}
\hline \hline
 & HellaSwag & OpenbookQA \\ \hline
C4 (num=10) & 63.374 & 36.600 \\
WikiText2 (num=10) & 63.730 & 37.200 \\
BookCorpus (num=10) & 62.179 & 36.600 \\
BookCorpus (num=20) & 62.420 & 36.200 \\
\hline \hline
\end{tabular}}
\caption{Performance impact of calibration dataset selection on LLaMA2-7B model. HellaSwag evaluates contextual understanding while OpenbookQA tests factual knowledge. }
\label{calibration}
\end{table}

\section{Conclusion}

By analyzing the correlation between outputs of different layers in the reproducing kernel Hilbert space, this paper reveals the ``patch-like'' relational patterns between layers in LLMs. Based on this insight, we propose Sliding-Window Merging method to compact the patch-redundant layers. This method dynamically merges consecutive layers using a similarity threshold, enabling rapid model compression while effectively preserving performance. Experimental results show that our method significantly outperforms existing pruning methods on both inference efficiency and zero-shot tasks. Moreover, our method can be seamlessly integrated with width pruning methods, leading to pruned models that achieve enhanced performance. We hope this study will inspire further research into depth-wise pruning methods and foster the development of a unified framework that combines both depth-wise and width-wise pruning strategies, ultimately contributing to the efficient deployment of LLMs in resource-constrained environments.

\bibliography{aaai2026}

\end{document}


\maketitle

\appendix

\section{Experimental Setup}

\subsection{Baseline Methods}

We primarily consider the width-wise pruning methods and depth-wise pruning methods as our baseline methods in our analysis. The specific information of baseline methods are described below, where we use their official code for implementation. To ensure a fair comparison, we employ the same calibration dataset across all methods. 

\textbf{Width-wise method.} The width-wise pruning methods include Wanda-sp, FLAP and LLM-Pruner.  

Wanda-sp is a variant of Wanda \citep{sun2024transformer}. The original metric of Wanda was based on the product of weight magnitudes and input activation norms, while Wanda-sp presented in \citep{an2024fluctuation} extends this in a struced way while using common dimensions among different modules. 

\begin{table}[h]
\centering
\renewcommand{\arraystretch}{1.25}
\resizebox{1.0\columnwidth}{!}{%
\begin{tabular}{c|cccc}
\hline \hline
Model & Pruned Ratio & Metrics & Structure & Params \\ \hline
\multirow{2}{*}{LLaMA2-7B / Vicuna-7B} & 0.2 & WIFV & AL-AM & 5470556160 \\
 & 0.35 & WIFV & AL-AM & 4485812224 \\ \hline
\multirow{2}{*}{LLaMA2-13B / Vicuna-13B} & 0.2 & WIFV & AL-AM & 10479211520 \\
 & 0.35 & WIFV & AL-AM & 8575800320 \\ \hline \hline
\end{tabular}}
\caption{Hyperparameter settings for Wanda-sp.}
\label{wanda-sp}
\end{table}

FLAP \citep{an2024fluctuation} is a new LLM retraining-free structured pruning framework that determines the recoverability of the feature map after removing weight columns based on the fluctuation pruning index. It adaptively determines the compressed model structure using normalized importance scores and adds additional bias terms to the pruned feature maps to restore performance.

\begin{table}[h]
\centering
\renewcommand{\arraystretch}{1.25}
\resizebox{1.0\columnwidth}{!}{%
\begin{tabular}{c|cccc}
\hline \hline
Model & Pruned Ratio & Metrics & Structure & Params \\ \hline
\multirow{2}{*}{LLaMA2-7B / Vicuna-7B} & 0.2 & WIFV & AL-AM & 5444964352 \\
 & 0.35 & WIFV & AL-AM & 4473626624 \\ \hline
\multirow{2}{*}{LLaMA2-13B / Vicuna-13B} & 0.2 & WIFV & AL-AM & 10481551360 \\
 & 0.35 & WIFV & AL-AM & 8578908160 \\ \hline
\multirow{2}{*}{Meta-LLaMA3} & 0.2 & WIFV & AL-AM & 6721589248 \\
 & 0.35 & WIFV & AL-AM & 5631029248 \\ \hline \hline
\end{tabular}}
\caption{Hyperparameter settings for FLAP.}
\label{flap}
\end{table}

LLM-Pruner \citep{ma2023llm} utilizes a Taylor-based importance metric to identify and remove attention heads from MHA and intermediate neurons from FFN. The pruning process is conducted locally, selecting removable groups within each module while ensuring that the dimensions across the blocks remain consistent. Following the original approach, we preserve the first and last few blocks without pruning. Both their pruned models and ours are retrained using Low-Rank Adaptation (LoRA).

\begin{table}[h]
\centering
\renewcommand{\arraystretch}{1.25}
\resizebox{1.0\columnwidth}{!}{%
\begin{tabular}{c|cccc}
\hline \hline
Model & Pruned Ratio & \multicolumn{1}{l}{\begin{tabular}[c]{@{}l@{}}Block\_mlp\_layer\_start /\\ BloCK\_attention\_layer\_start\end{tabular}} & \multicolumn{1}{l}{\begin{tabular}[c]{@{}l@{}}Block\_mlp\_layer\_end /\\ Block\_attention\_layer\_start\end{tabular}} & Params \\ \hline
\multirow{2}{*}{LLaMA2-7B/Vicuna-7B} & 0.24 & 4 & 30 & 5512646656 \\
 & 0.43 & 4 & 30 & 4517122048 \\ \hline
\multirow{2}{*}{LLaMA2-13B/Vicuna-13B} & 0.24 & 4 & 38 & 10480911360 \\
 & 0.42 & 4 & 38 & 8557153280 \\ \hline
\multirow{2}{*}{LLaMA3-8B} & 0.24 & 4 & 30 & 6794588160 \\
 & 0.43 & 4 & 30 & 5651673088 \\ \hline \hline
\end{tabular}}
\caption{Hyperparameter settings for LLM-Pruner.}
\label{llm-pruner}
\end{table}

\textbf{Depth-wise method.} The depth-wise pruning methods use the Transformer module in LLM as the pruning unit. SLEB \citep{song2024sleb} uses a logit-based approach to find unnecessary blocks and updates the importance score after removing each block. SLEB pursues a no-retraining setting, but it cannot maintain sufficient performance as the pruning rate increases. Shortened-LLM \citep{kim2024shortened} uses the PPL standard to determine the importance of the transformer layer, and deletes unimportant transformer layers after sorting. Shortened-LLM method uses lora to retrain to restore the performance of the model.

\begin{table}[h]
\centering
\renewcommand{\arraystretch}{1.25}
\resizebox{1.0\columnwidth}{!}{%
\begin{tabular}{c|ccccc}
\hline \hline
Model & Pruned Ratio & Block & Head & FFN-D & Params \\ \hline
\multirow{2}{*}{LLaMA2-7B/Vicuna-7B} & 0.2 & 26 & 32 & 11008 & 5524115456 \\
 & 0.35 & 21 & 32 & 11008 & 4512198656 \\ \hline
\multirow{2}{*}{LLaMA2-13B/Vicuna-13B} & 0.2 & 32 & 40 & 13824 & 10478228480 \\
 & 0.35 & 26 & 40 & 13824 & 8575001600 \\ \hline
\multirow{2}{*}{LLaMA3-8B} & 0.2 & 26 & 32 & 14336 & 6721589248 \\
 & 0.35 & 21 & 32 & 14336 & 5631029248 \\ \hline \hline
\end{tabular}}
\caption{Hyperparameter settings for depth-wise methods.}
\label{depth-wise}
\end{table}

\subsection{Selected Transformer Layers}
We summarize the number and indices of transformer blocks selected for removal using our method in Tab.\ref{selected}. The specified sets of Transformer layers are fused into a single layer, represented by the index of the first layer in the set (e.g., [25, 26, 27, 28, 29, 30] are fused into layer 25).

\begin{table}[H]
\centering
\renewcommand{\arraystretch}{1.3}
\resizebox{1.0\columnwidth}{!}{%
\begin{tabular}{c|c|c|c}
\hline \hline
Model & Threshold & Num of layers & Merged Layers \\ \hline
\multirow{2}{*}{LLaMA2-7B} & 0.81 & 26 & {[}{[}25, 26, 27, 28, 29, 30{]}, {[}10, 11{]}{]} \\
 & 0.68 & 21 & {[}{[}22, 23, 24, 25, 26, 27, 28, 29, 30{]}, {[}14, 15{]}, {[}10, 11{]}, {[}6,   7{]}{]} \\
\multirow{2}{*}{Vicuna-7B} & 0.78 & 26 & {[}{[}25, 26, 27, 28, 29, 30{]}, {[}13, 14{]}{]} \\
 & 0.57 & 21 & {[}{[}21, 22, 23, 24, 25, 26, 27, 28, 29, 30{]}, {[}13, 14{]}, {[}10, 11{]}{]} \\
\multirow{2}{*}{LLaMA3-8B} & 0.72 & 26 & {[}{[}28, 29, 30{]}, {[}25, 26, 27{]}, {[}22, 23{]}, {[}14, 15{]}{]} \\
 & 0.51 & 21 & {[}{[}25, 26, 27, 28, 29, 30{]}, {[}21, 22{]}, {[}19, 20{]}, {[}14, 15, 16{]},   {[}12, 13{]}, {[}10, 11{]}{]} \\
\multirow{2}{*}{LLaMA2-13B} & 0.8 & 32 & {[}{[}30, 31, 32, 33, 34, 35, 36, 37, 38{]}{]} \\
 & 0.72 & 26 & {[}{[}26, 27, 28, 29, 30, 31, 32, 33, 34, 35, 36, 37, 38{]}, {[}16,   17{]}, {[}12, 13{]}{]} \\
\multirow{2}{*}{Vicuna-13B} & 0.76 & 32 & {[}{[}33, 34, 35, 36, 37, 38{]}, {[}31, 32{]}, {[}29, 30{]}, {[}15, 16{]}{]} \\
 & 0.65 & 26 & {[}{[}31, 32, 33, 34, 35, 36, 37, 38{]}, {[}29, 30{]}, {[}27, 28{]}, {[}9, 10,   11, 12, 13, 14{]}{]} \\ \hline \hline
\end{tabular}}
\caption{Corresponding layer indices of the pruned models under different threshold setting of our depth-wise pruning method used in the main results}
\label{selected}
\end{table}

\subsection{LoRA Retraining}
We apply a LoRA adapter \citep{hulora} to every projection weight matrix by following \citep{ma2023llm}. We employ a LoRA rank of 8, a learning rate of 0.0001, and a batch size of 64 over 2 epochs. The retraining costs are notably low, with the entire process being executed on a single NVIDIA A100 (80GB VRAM) GPU. For example, retraining a 20\%-pruned model from 7B parameters takes about 2 hours and utilizes 22GB GPU memory. Here we provide a brief introducion of LoRA retraining to ensure the self-contained aspect of our work. 

Each learnable weight matrix (including both pruned and unpruned linear projections in LLMs) is represented by $W$. The update to $W$, denoted as $\triangle W$, is factorized as $\triangle W = PQ \in R^{d^{-} \times d^+}$, where $P \in R^{d^{-} \times d}$ and $Q \in R^{d \times d^+}$. Here, $d^{-}$, $d$, and $d^+$ correspond to the dimensions of the input, hidden, and output layers, respectively. The forward computation is then expressed as follows:
\begin{equation}
f(x) = (W+\triangle W)x + b =(WX+b)+(PQ)X,
\end{equation}
where $b$ represents the bias in the dense layer. By training only the low-rank matrices $P$ and $Q$, we considerably reduce both the computational complexity and the dependence on large-scale training data. Furthermore, the additional parameters $P$ and $Q$ can be reparameterized as $\triangle W$, thereby introducing no extra parameters in the final pruned model.

\section{Supplementary Experiment Results}

\subsection{Additional Results of Inference Efficiency}
Tab.\ref{efficiency} shows the inference of our depth-wise pruning method in different LLMs (LLaMA2-7B, LLaMA2-13B, Vicuna-13B-v1.3, and LLaMA3-8B). As the pruning ratio increases, the latency of the model decreases, the throughput increases, and the GPU memory usage and number of parameters also decrease accordingly.

\begin{table}[h]
\centering
\renewcommand{\arraystretch}{1.15}
\resizebox{1.0\columnwidth}{!}{%
\begin{tabular}{c|cccc}
\hline \hline
 & Latency(sec) & Throughout(tokens/s) & GPU\_Memory & nparam \\ \hline 
LLaMA2-7B & 2.729 & 46.905 & 13020.25 & 6738415616 \\ 
20\% Pruned & 2.339 & 54.758 & 10682.45 & 5524115456 \\
35\% Pruned & 1.889 & 67.770 & 8725.9 & 4512198656 \\ \hline 
LLaMA2-13B & 3.635 & 35.210 & 25188.85 & 13015864320 \\ 
20\% Pruned & 2.908 & 44.016 & 20274.25 & 10478228480 \\
35\% Pruned & 2.380 & 53.793 & 16593.1 & 8575001600 \\ \hline 
Vicuna-7B-v1.3 & 2.865 & 44.681 & 13021.8 & 6738415616 \\
20\% Pruned & 2.370 & 54.065 & 10682.45 & 5524115456 \\
35\% Pruned & 2.000 & 64.087 & 8725.9 & 4512198656 \\ \hline 
Vicuna-13B-v1.3 & 3.593 & 35.623 & 25186.95 & 13015864320 \\ 
20\% Pruned & 2.885 & 44.386 & 20274.25 & 10478228480 \\
35\% Pruned & 2.395 & 53.461 & 16593.1 & 8575001600 \\ \hline 
LLaMA3-8B & 3.150 & 40.640 & 15364 & 8030261248 \\ 
20\% Pruned & 2.635 & 48.616 & 12862 & 6721589248 \\
35\% Pruned & 2.218 & 57.795 & 10776 & 5631029248 \\ \hline \hline
\end{tabular}}
\caption{Inference efficiency of our pruned models. (Measured with 12 input tokens, 128 output tokens, and a batch size of 1 on a NVIDIA H100 GPU.)}
\label{efficiency}
\end{table}

\begin{table}[h]
\centering
\renewcommand{\arraystretch}{1.25}
\resizebox{1.0\columnwidth}{!}{%
\begin{tabular}{ccc|cc|c|c}
\hline \hline
\multicolumn{3}{c|}{Depth-wise(ours)} & \multicolumn{2}{c|}{Width-wise(LLM-Pruner)} & \multirow{2}{*}{Nparam} & \multirow{2}{*}{AVE} \\Proportion & Threshold & Remove layers & Proportion & Pruner\_ratio &  &  \\ \hline
0\% &  & 0 & 100\% & 0.43 & 4532772064 & 52.366 \\
18\% & 0.96 & 2 & 82\% & 0.35 & 4502876160 & 54.372 \\
36\% & 0.86 & 4 & 64\% & 0.3 & 4501803008 & 55.753 \\
53\% & 0.81 & 6 & 47\% & 0.24 & 4486926336 & 55.188 \\
71\% & 0.77 & 8 & 29\% & 0.16 & 4476604416 & 53.967 \\
91\% & 0.7 & 10 & 9\% & 0.06 & 4530630656 & 52.780 \\
100\% & 0.68 & 11 & 0\% & & 4512198656 & 51.869 \\ \hline \hline
\end{tabular}}
\caption{Pruning proportions and corresponding parameter settings of integrated method on the LLaMA2-7B model.}
\label{integrate}
\end{table}

\subsection{Integration of Depth-wise Pruning and Width-wise Pruning}
We integrate our method with LLM-Pruner to further improve the pruning effect. In the first stage, we apply our proposed pruning method, based on the layer compression model. In the second stage, we use the LLM-Pruner method to remove unnecessary coupling structures from the model obtained in the first stage. As a result, the pruned model achieved a pruning rate of 35\% relative to the original model. Tab.\ref{integrate} shows the pruning proportions and corresponding parameter settings of depth-wise pruning and width-wise pruning on the LLaMA2-7B model using the integrated method.

\begin{figure}[h]
\centering
\includegraphics[width=1.0\columnwidth]{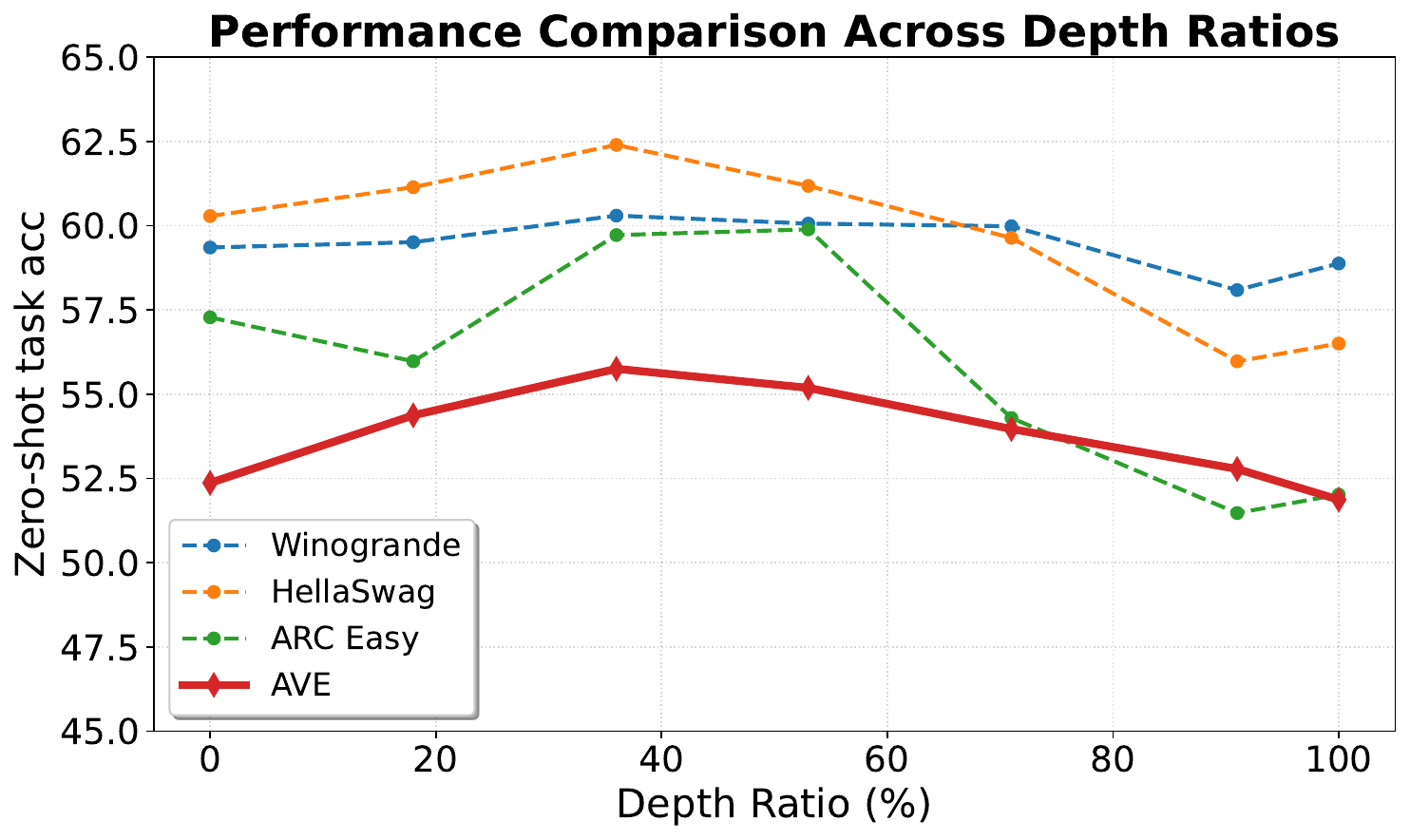}  
\caption{Performance of the integrated method on the Vicuna-13B model.}
\label{merge2}
\end{figure}

We also tested our integrated method on the Vicuna-13B model, and the experimental results are shown in Fig\ref{merge2}. Tab.\ref{integrate2}  shows the pruning proportions and corresponding parameter settings of depth-wise pruning and width-wise pruning.

\begin{table}[h]
\centering
\renewcommand{\arraystretch}{1.25}
\resizebox{1.0\columnwidth}{!}{%
\begin{tabular}{ccc|cc|c|c}
\hline \hline
\multicolumn{3}{c|}{Depth-wise(ours)} & \multicolumn{2}{c|}{Width-wise(LLM-Pruner)} & \multirow{2}{*}{Nparam} & \multirow{2}{*}{AVE} \\
Proportion & Threshold & Remove layers & Proportion & Pruner\_ratio &  &  \\ \hline
0\% &  & 0 & 100\% & 0.42 & 8557153280 & 59.863 \\
14\% & 0.94 & 2 & 86\% & 0.35 & 8606530560 & 61.236 \\
28\% & 0.86 & 4 & 72\% & 0.3 & 8565928960 & 60.321 \\
43\% & 0.8 & 6 & 57\% & 0.28 & 8606039040 & 60.455 \\
57\% & 0.76 & 8 & 43\% & 0.24 & 8582722560 & 60.257 \\
71\% & 0.7 & 10 & 29\% & 0.16 & 8551490560 & 61.203 \\
100\% & 0.57 & 14 & 0\% &  & 8575001600 & 59.617 \\ \hline \hline
\end{tabular}}
\caption{Pruning proportions and corresponding parameter settings of integrated method on the Vicuna-13B model.}
\label{integrate2}
\end{table}

\subsection{Three different layer merging methods}
For the $l$-th layer of an LLM, we denote all its parameters as $\theta_l$. We considered three different layer merging methods to merge the parameters of the subsequent m consecutive layers $\theta_{l+1},\theta_{l+2},...,\theta_{l+m}$ into $\theta_l$ to form $ \theta_l^*$. 

\begin{itemize}
\item \textbf{Remove:} We directly ignore the parameters of the $l+1$ to $l+m$ layers, and only retain the parameters of the $l$-th layer (although in actual operation, we may still label the combined parameter set as $\theta_{l}^*$, but here $\theta_{l}^*$ actually only contains the parameters of the original $l$-th layer).
\begin{equation}
    \theta_l^* = \theta_l    
\end{equation}
\item \textbf{Average:} We calculate the average value of the parameters from layer $l$ to $l+m$, and use this average value to form a new parameter set $\theta_{l}^*$.
\begin{equation}
    \theta_l^* = \frac{\sum_{i=l}^{l+m}\theta_i}{m}    
\end{equation}
\item \textbf{Our method:} We adopt a parameter merging strategy based on inter-layer differences, adding the differences between adjacent layer and base layer parameters to gradually integrate redundant information.
\begin{equation}
    \theta_l^* = \theta_l + \sum_{i=l+1}^{l+m}(\theta_i-\theta_l)   
\end{equation}

\end{itemize}

\begin{table}[h]
\centering
\renewcommand{\arraystretch}{1.2}
\resizebox{1.0\columnwidth}{!}{%
\begin{tabular}{ccc}
\hline \hline
Threshold & Num of layers & Merged Layers \\ \hline
0.95 & 30 & {[}{[}27, 28{]}, {[}13, 14{]}{]} \\
0.9 & 30 & {[}{[}27, 28, 29{]}{]} \\
0.85 & 29 & {[}{[}29, 30{]}, {[}27, 28{]}, {[}24, 25{]}{]} \\
0.8 & 29 & {[}{[}27, 28, 29, 30{]}{]} \\
0.75 & 26 & {[}{[}26, 27, 28, 29, 30{]}, {[}24, 25{]}, {[}9, 10{]}{]} \\
0.7 & 24 & {[}{[}24, 25, 26, 27, 28, 29, 30{]}, {[}22, 23{]}, {[}9, 10{]}{]} \\
0.65 & 22 & {[}{[}22, 23, 24, 25, 26, 27, 28, 29, 30{]}, {[}17, 18{]}, {[}9, 10{]}{]} \\ \hline \hline
\end{tabular}}
\caption{The layer index corresponding to the pruned model obtained by the "delete" layer merging method under different threshold settings.}
\label{delete}
\end{table}

\begin{table}[h]
\centering
\renewcommand{\arraystretch}{1.2}
\resizebox{1.0\columnwidth}{!}{%
\begin{tabular}{ccc}
\hline \hline
Threshold & Num of layers & Merged Layers \\ \hline
0.95 & 31 & {[}{[}21, 22{]}{]} \\
0.9 & 31 & {[}{[}28, 29{]}{]} \\
0.85 & 31 & {[}{[}29, 30{]}{]} \\
0.8 & 30 & {[}{[}29, 30{]}, {[}23, 24{]}{]} \\
0.75 & 29 & {[}{[}29, 30{]}, {[}26, 27{]}, {[}22,   23{]}{]} \\
0.7 & 28 & {[}{[}29, 30{]}, {[}27, 28{]}, {[}23,   24{]}, {[}10, 11{]}{]} \\
0.65 & 27 & {[}{[}29, 30{]}, {[}27, 28{]}, {[}25,   26{]}, {[}22, 23{]}, {[}10, 11{]}{]} \\ \hline \hline
\end{tabular}}
\caption{The layer index corresponding to the pruned model obtained by the "average" layer merging method under different threshold settings.}
\label{average}
\end{table}

\subsection{Zero-shot performance in larger scale}
Tab.\ref{large-scale} presents the zero-shot performance of various downstream tasks with the proposed method applied to the LLaMA2-13B model and Vicuna-13B model. Our method shows superior pruning capabilities.

\begin{table*}[h]
\centering
\renewcommand{\arraystretch}{1.25}
\resizebox{1.9\columnwidth}{!}{%
\begin{tabular}{ccc|llllllll}
\hline \hline
\multicolumn{3}{c|}{\#Param \& Method} & \multicolumn{1}{c}{BoolQ} & \multicolumn{1}{c}{PIQA} & \multicolumn{1}{c}{HellaSwag} & \multicolumn{1}{c}{WinoGrande} & \multicolumn{1}{c}{ARC-easy} & \multicolumn{1}{c}{ARC-challenge} & \multicolumn{1}{c}{OpenbookQA} & \multicolumn{1}{c}{AVE} \\ \hline \hline
\multicolumn{3}{c|}{LLaMA-2-13B(Original)} & 80.550 & 79.053 & 79.367 & 72.139 & 79.377 & 49.147 & 45.200 & 69.262 \\ \hline 
\multicolumn{1}{c|}{} & \multicolumn{1}{c|}{} & Wanda-sp & 69.630 & 77.480 & 74.750 & 67.010 & 73.480 & 44.110 & 44.000 & 64.351 \\
\multicolumn{1}{c|}{} & \multicolumn{1}{c|}{} & FLAP & 72.780 & 74.650 & 69.070 & 68.350 & 70.830 & 40.610 & 40.000 & 62.327 \\
\multicolumn{1}{c|}{} & \multicolumn{1}{c|}{\multirow{-3}{*}{width}} & LLM-Pruner & 71.315 & 79.162 & 74.836 & 67.324 & 73.485 & 43.771 & 41.600 & 64.499 \\ \cline{2-11} 
\multicolumn{1}{c|}{} & \multicolumn{1}{c|}{} & SLEB & 63.211 & 76.061 & 70.116 & 65.430 & 70.749 & 39.932 & 39.200 & 60.671 \\
\multicolumn{1}{c|}{} & \multicolumn{1}{c|}{} & Shortened-LLM & 68.318 & 76.279 & 75.204 & 71.113 & 74.790 & 46.672 & 42.400 & 64.968 \\
\multicolumn{1}{c|}{\multirow{-6}{*}{20\%Pruned}} & \multicolumn{1}{c|}{\multirow{-3}{*}{depth}} & Ours & 64.006 & 77.421 & 76.369 & 71.665 & 76.557 & 48.549 & 43.800 & \textbf{65.481} \\ \hline
\multicolumn{1}{c|}{} & \multicolumn{1}{c|}{} & Wanda-sp & 59.020 & 55.110 & 33.580 & 52.800 & 29.840 & 24.910 & 28.600 & 40.551 \\
\multicolumn{1}{c|}{} & \multicolumn{1}{c|}{} & FLAP & 71.250 & 69.590 & 61.680 & 64.400 & 59.050 & 34.130 & 36.000 & 56.586 \\
\multicolumn{1}{c|}{} & \multicolumn{1}{c|}{\multirow{-3}{*}{width}} & LLM-Pruner & 66.086 & 76.061 & 67.785 & 59.195 & 67.382 & 39.334 & 41.800 & 59.663 \\ \cline{2-11} 
\multicolumn{1}{c|}{} & \multicolumn{1}{c|}{} & SLEB & 62.385 & 70.620 & 58.415 & 55.643 & 62.500 & 36.689 & 33.800 & 54.293 \\
\multicolumn{1}{c|}{} & \multicolumn{1}{c|}{} & Shortened-LLM & 63.333 & 72.307 & 67.128 & 63.694 & 66.667 & 39.761 & 37.000 & 58.556 \\
\multicolumn{1}{c|}{\multirow{-6}{*}{35\%Pruned}} & \multicolumn{1}{c|}{\multirow{-3}{*}{depth}} & Ours & 59.939 & 73.286 & 68.751 & 65.983 & 68.981 & 41.724 & 39.000 & \textbf{59.666} \\ \hline \hline
\multicolumn{3}{c|}{Vicuna-13B-v1.3(Original)} & 82.813 & 78.346 & 77.017 & 71.113 & 75.547 & 47.611 & 45.400 & 68.264 \\ \hline
\multicolumn{1}{c|}{} & \multicolumn{1}{c|}{} & Wanda-sp & 77.090 & 77.090 & 74.420 & 67.960 & 67.800 & 42.320 & 42.800 & 64.211 \\
\multicolumn{1}{c|}{} & \multicolumn{1}{c|}{} & FLAP & 81.100 & 76.770 & 73.720 & 68.350 & 71.510 & 42.490 & 41.000 & 64.991 \\
\multicolumn{1}{c|}{} & \multicolumn{1}{c|}{\multirow{-3}{*}{width}} & LLM-Pruner & 74.526 & 78.346 & 72.426 & 69.219 & 69.739 & 40.529 & 43.200 & 63.998 \\ \cline{2-11} 
\multicolumn{1}{c|}{} & \multicolumn{1}{c|}{} & SLEB & 62.385 & 70.620 & 58.415 & 55.643 & 62.500 & 36.689 & 33.800 & 54.293 \\
\multicolumn{1}{c|}{} & \multicolumn{1}{c|}{} & Shortened-LLM & 75.535 & 77.476 & 73.571 & 68.272 & 72.180 & 44.198 & 43.200 & 64.919 \\
\multicolumn{1}{c|}{\multirow{-6}{*}{20\%Pruned}} & \multicolumn{1}{c|}{\multirow{-3}{*}{depth}} & Ours & 81.040 & 76.442 & 74.846 & 70.324 & 71.801 & 44.625 & 41.800 & \textbf{65.840} \\ \hline
\multicolumn{1}{c|}{} & \multicolumn{1}{c|}{} & Wanda-sp & 61.650 & 71.220 & 63.960 & 61.400 & 57.490 & 35.320 & 37.000 & 55.434 \\
\multicolumn{1}{c|}{} & \multicolumn{1}{c|}{} & FLAP & 75.170 & 73.990 & 65.540 & 67.560 & 61.320 & 36.770 & 37.400 & 59.679 \\
\multicolumn{1}{c|}{} & \multicolumn{1}{c|}{\multirow{-3}{*}{width}} & LLM-Pruner & 70.581 & 76.659 & 67.317 & 65.272 & 63.258 & 35.154 & 40.800 & 59.863 \\ \cline{2-11} 
\multicolumn{1}{c|}{} & \multicolumn{1}{c|}{} & SLEB & 37.829 & 51.034 & 25.503 & 50.908 & 26.221 & 27.218 & 27.400 & 35.159 \\
\multicolumn{1}{c|}{} & \multicolumn{1}{c|}{} & Shortened-LLM & 68.532 & 74.102 & 66.421 & 64.009 & 67.593 & 41.126 & 38.600 & \textbf{60.055} \\
\multicolumn{1}{c|}{\multirow{-6}{*}{35\%Pruned}} & \multicolumn{1}{c|}{\multirow{-3}{*}{depth}} & Ours & 69.450 & 74.918 & 67.447 & 63.378 & 66.414 & 38.311 & 37.400 & 59.617 \\ \hline \hline
\end{tabular}
}
\caption{Performance comparison of pruning methods across multiple baselines on LLaMA2-13B and Vicuna-13B models.}
\label{large-scale}
\end{table*}

\subsection{Additional Results of Moderate Pruning and LoRA Retraining}

Tab.\ref{lora1}-\ref{lora5} show the zero-shot results of several pruning strategies that require retraining, including LLM-Pruner, Shortened-LLM, and our proposed method on different model.

\begin{table*}[h]
\centering
\renewcommand{\arraystretch}{1.25}
\resizebox{2.0\columnwidth}{!}{%
\begin{tabular}{cl|cccccccc}
\hline \hline
\multicolumn{2}{c|}{20\%Pruned} & BoolQ & PIQA & HellaSwag & WinoGrande & ARC-easy & ARC-challenge & OpenbookQA & AVE \\ \hline 
\multirow{2}{*}{LLM\_Pruner} & wo\_lora & 53.761 & 76.659 & 66.132 & 61.168 & 64.857 & 37.884 & 40.200 & 57.237 \\
 & w\_lora & 63.731 & 77.476 & 67.128 & 61.878 & 65.783 & 38.481 & 40.400 & 59.268 \\
\multirow{2}{*}{Shortened-LLM} & wo\_lora & 60.489 & 73.776 & 63.364 & 57.459 & 64.015 & 33.191 & 36.200 & 55.499 \\
 & w\_lora & 61.560 & 76.061 & 67.994 & 58.800 & 68.813 & 37.884 & 38.000 & 58.445 \\
\multirow{2}{*}{Ours} & wo\_lora & 62.324 & 70.239 & 65.097 & 66.298 & 61.448 & 38.311 & 37.800 & 57.360 \\
 & w\_lora & 69.450 & 73.667 & 70.484 & 67.088 & 69.108 & 41.212 & 42.600 & 61.944 \\ \hline \hline
\multicolumn{2}{c|}{35\%Pruned} & BoolQ & PIQA & HellaSwag & WinoGrande & ARC-easy & ARC-challenge & OpenbookQA & AVE \\ \hline 
\multirow{2}{*}{LLM\_Pruner} & wo\_lora & 50.703 & 69.695 & 51.275 & 52.960 & 49.495 & 31.399 & 36.000 & 48.790 \\
 & \multicolumn{1}{c|}{w\_lora} & 45.260 & 74.755 & 60.287 & 59.353 & 57.281 & 32.423 & 37.200 & 52.366 \\
\multirow{2}{*}{Shortened-LLM} & wo\_lora & 61.101 & 67.791 & 45.987 & 52.960 & 48.485 & 27.218 & 32.800 & 48.049 \\
 & \multicolumn{1}{c|}{w\_lora} & 62.171 & 71.926 & 54.800 & 51.776 & 59.259 & 30.887 & 35.400 & 52.317 \\
\multirow{2}{*}{Ours} & wo\_lora & 62.171 & 63.874 & 52.131 & 59.511 & 46.170 & 31.826 & 32.600 & 49.755 \\
 & \multicolumn{1}{c|}{w\_lora} & 63.119 & 65.452 & 56.503 & 58.879 & 52.020 & 31.911 & 35.200 & 51.869 \\ \hline \hline
\end{tabular}}
\caption{Performance with/without LoRA retraining on LLaMA2-7B.}
\label{lora1}
\end{table*}

\begin{table*}[h]
\centering
\renewcommand{\arraystretch}{1.25}
\resizebox{2.0\columnwidth}{!}{%
\begin{tabular}{cl|cccccccc}
\hline \hline
\multicolumn{2}{c|}{20\%Pruned} & BoolQ & PIQA & HellaSwag & WinoGrande & ARC-easy & ARC-challenge & OpenbookQA & AVE \\ \hline 
\multirow{2}{*}{LLM\_Pruner} & wo\_lora & 57.554 & 74.810 & 65.216 & 59.037 & 64.815 & 36.263 & 39.400 & 56.728 \\
 & w\_lora & 67.645 & 76.115 & 66.660 & 63.931 & 65.446 & 36.604 & 40.400 & 59.543 \\
\multirow{2}{*}{Shortened-LLM} & wo\_lora & 63.670 & 72.470 & 63.105 & 62.194 & 64.352 & 36.775 & 35.600 & 56.881 \\
 & w\_lora & 72.355 & 74.701 & 67.576 & 64.562 & 70.034 & 38.225 & 38.600 & 60.865 \\
\multirow{2}{*}{Ours} & wo\_lora & 63.394 & 72.742 & 66.152 & 66.219 & 66.919 & 39.078 & 40.600 & 59.301 \\
 & w\_lora & 77.431 & 74.755 & 69.040 & 68.745 & 69.318 & 38.908 & 40.200 & 62.628 \\ \hline \hline
\multicolumn{2}{c|}{35\%Pruned} & BoolQ & PIQA & HellaSwag & WinoGrande & ARC-easy & ARC-challenge & OpenbookQA & AVE \\ \hline 
\multirow{2}{*}{LLM\_Pruner} & wo\_lora & 60.642 & 70.294 & 54.860 & 52.881 & 53.662 & 33.618 & 36.000 & 51.708 \\
 & \multicolumn{1}{c|}{w\_lora} & 63.976 & 73.069 & 59.560 & 58.564 & 56.524 & 32.679 & 37.800 & 54.596 \\
\multirow{2}{*}{Shortened-LLM} & wo\_lora & 60.245 & 64.527 & 43.856 & 53.986 & 47.348 & 26.792 & 29.600 & 46.622 \\
 & \multicolumn{1}{c|}{w\_lora} & 64.281 & 70.783 & 56.722 & 57.380 & 59.596 & 31.485 & 34.000 & 53.464 \\
\multirow{2}{*}{Ours} & wo\_lora & 62.202 & 66.431 & 53.834 & 61.484 & 52.946 & 33.532 & 33.400 & 51.976 \\
 & \multicolumn{1}{c|}{w\_lora} & 69.235 & 70.294 & 60.705 & 62.273 & 60.227 & 33.618 & 37.400 & 56.250 \\ \hline \hline
\end{tabular}}
\caption{Performance with/without LoRA retraining on Vicuna-7B-v1.3.}
\label{lora2}
\end{table*}

\begin{table*}[h]
\centering
\renewcommand{\arraystretch}{1.25}
\resizebox{2.0\columnwidth}{!}{%
\begin{tabular}{cl|cccccccc}
\hline \hline
\multicolumn{2}{c|}{20\%Pruned} & BoolQ & PIQA & HellaSwag & WinoGrande & ARC-easy & ARC-challenge & OpenbookQA & AVE \\ \hline 
\multirow{2}{*}{LLM\_Pruner} & wo\_lora & 56.942 & 77.040 & 67.785 & 68.666 & 68.603 & 39.078 & 40.400 & 59.788 \\
 & w\_lora & 74.037 & 79.489 & 74.268 & 70.324 & 74.285 & 46.587 & 42.600 & 65.941 \\
\multirow{2}{*}{Shortened-LLM} & wo\_lora & 45.443 & 73.232 & 60.994 & 57.853 & 65.404 & 34.044 & 36.000 & 53.282 \\
 & w\_lora & 66.208 & 78.074 & 72.695 & 62.747 & 75.295 & 44.795 & 43.400 & 63.316 \\
\multirow{2}{*}{Ours} & wo\_lora & 38.073 & 71.980 & 61.800 & 69.613 & 66.035 & 41.809 & 38.400 & 55.387 \\
 & w\_lora & 76.789 & 77.639 & 73.770 & 71.744 & 76.599 & 50.939 & 41.200 & 66.954 \\ \hline \hline
\multicolumn{2}{c|}{35\%Pruned} & BoolQ & PIQA & HellaSwag & WinoGrande & ARC-easy & ARC-challenge & OpenbookQA & AVE \\ \hline 
\multirow{2}{*}{LLM\_Pruner} & wo\_lora & 47.829 & 69.369 & 45.150 & 53.118 & 48.485 & 27.816 & 33.200 & 46.424 \\
 & \multicolumn{1}{c|}{w\_lora} & 64.465 & 74.048 & 61.800 & 59.353 & 64.646 & 34.386 & 37.200 & 56.557 \\
\multirow{2}{*}{Shortened-LLM} & wo\_lora & 61.651 & 66.431 & 49.801 & 51.697 & 51.431 & 29.352 & 30.400 & 48.680 \\
 & \multicolumn{1}{c|}{w\_lora} & 63.180 & 72.851 & 62.985 & 58.090 & 66.877 & 37.116 & 37.000 & 56.871 \\
\multirow{2}{*}{Ours} & wo\_lora & 40.428 & 62.350 & 40.291 & 55.485 & 39.394 & 28.242 & 28.200 & 42.056 \\
 & \multicolumn{1}{c|}{w\_lora} & 67.554 & 73.830 & 61.472 & 62.747 & 64.352 & 36.007 & 37.600 & 57.652 \\ \hline \hline
\end{tabular}}
\caption{Performance with/without LoRA retraining on LLaMA3-8B.}
\label{lora3}
\end{table*}

\begin{table*}[h]
\centering
\renewcommand{\arraystretch}{1.25}
\resizebox{2.0\columnwidth}{!}{%
\begin{tabular}{cl|cccccccc}
\hline \hline
\multicolumn{2}{c|}{20\%Pruned} & BoolQ & PIQA & HellaSwag & WinoGrande & ARC-easy & ARC-challenge & OpenbookQA & AVE \\ \hline 
\multirow{2}{*}{LLM\_Pruner} & wo\_lora & 65.566 & 78.509 & 72.018 & 64.167 & 69.992 & 43.601 & 40.600 & 62.065 \\
 & w\_lora & 71.315 & 79.162 & 74.836 & 67.324 & 73.485 & 43.771 & 41.600 & 64.499 \\
\multirow{2}{*}{Shortened-LLM} & wo\_lora & 63.180 & 75.027 & 71.191 & 70.481 & 69.529 & 43.089 & 40.800 & 61.900 \\
 & w\_lora & 68.318 & 76.279 & 75.204 & 71.113 & 74.790 & 46.672 & 42.400 & 64.968 \\
\multirow{2}{*}{Ours} & wo\_lora & 38.318 & 72.361 & 67.726 & 70.797 & 64.815 & 39.676 & 42.800 & 56.642 \\
 & w\_lora & 64.006 & 77.421 & 76.369 & 71.665 & 76.557 & 48.549 & 43.800 & 65.481 \\ \hline \hline
\multicolumn{2}{c|}{35\%Pruned} & BoolQ & PIQA & HellaSwag & WinoGrande & ARC-easy & ARC-challenge & OpenbookQA & AVE \\ \hline 
\multirow{2}{*}{LLM\_Pruner} & wo\_lora & 52.049 & 74.048 & 61.522 & 55.801 & 60.816 & 36.689 & 39.800 & 54.389 \\
 & \multicolumn{1}{c|}{w\_lora} & 66.086 & 76.061 & 67.785 & 59.195 & 67.382 & 39.334 & 41.800 & 59.663 \\
\multirow{2}{*}{Shortened-LLM} & wo\_lora & 62.141 & 69.042 & 59.361 & 60.221 & 53.830 & 31.741 & 34.400 & 52.962 \\
 & \multicolumn{1}{c|}{w\_lora} & 63.333 & 72.307 & 67.128 & 63.694 & 66.667 & 39.761 & 37.000 & 58.556 \\
\multirow{2}{*}{Ours} & wo\_lora & 40.489 & 67.900 & 56.234 & 63.694 & 51.347 & 33.532 & 36.400 & 49.942 \\
 & \multicolumn{1}{c|}{w\_lora} & 59.939 & 73.286 & 68.751 & 65.983 & 68.981 & 41.724 & 39.000 & 59.666 \\ \hline \hline
\end{tabular}}
\caption{Performance with/without LoRA retraining on LLaMA2-13B.}
\label{lora4}
\end{table*}

\begin{table*}[h]
\centering
\renewcommand{\arraystretch}{1.25}
\resizebox{2.0\columnwidth}{!}{%
\begin{tabular}{cl|cccccccc}
\hline \hline
\multicolumn{2}{c|}{20\%Pruned} & BoolQ & PIQA & HellaSwag & WinoGrande & ARC-easy & ARC-challenge & OpenbookQA & AVE \\ \hline 
\multirow{2}{*}{LLM\_Pruner} & wo\_lora & 74.006 & 77.040 & 71.679 & 64.799 & 66.498 & 39.078 & 41.800 & 62.129 \\
 & w\_lora & 74.526 & 78.346 & 72.426 & 69.219 & 69.739 & 40.529 & 43.200 & 63.998 \\
\multirow{2}{*}{Shortened-LLM} & wo\_lora & 66.239 & 74.918 & 70.056 & 66.456 & 68.476 & 43.857 & 39.600 & 61.372 \\
 & w\_lora & 75.535 & 77.476 & 73.571 & 68.272 & 72.180 & 44.198 & 43.200 & 64.919 \\
\multirow{2}{*}{Ours} & wo\_lora & 75.199 & 75.898 & 71.679 & 69.692 & 70.370 & 43.942 & 43.200 & 64.283 \\
 & w\_lora & 81.040 & 76.442 & 74.846 & 70.324 & 71.801 & 44.625 & 41.800 & 65.840 \\ \hline \hline
\multicolumn{2}{c|}{35\%Pruned} & BoolQ & PIQA & HellaSwag & WinoGrande & ARC-easy & ARC-challenge & OpenbookQA & AVE \\ \hline 
\multirow{2}{*}{LLM\_Pruner} & wo\_lora & 63.609 & 73.449 & 63.683 & 58.800 & 52.778 & 34.471 & 38.200 & 54.999 \\
 & \multicolumn{1}{c|}{w\_lora} & 70.581 & 76.659 & 67.317 & 65.272 & 63.258 & 35.154 & 40.800 & 59.863 \\
\multirow{2}{*}{Shortened-LLM} & wo\_lora & 42.385 & 69.532 & 57.897 & 60.063 & 60.017 & 37.372 & 34.800 & 51.724 \\
 & \multicolumn{1}{c|}{w\_lora} & 68.532 & 74.102 & 66.421 & 64.009 & 67.593 & 41.126 & 38.600 & 60.055 \\
\multirow{2}{*}{Ours} & wo\_lora & 63.609 & 73.123 & 58.703 & 61.089 & 62.837 & 36.604 & 34.600 & 55.795 \\
 & \multicolumn{1}{c|}{w\_lora} & 69.450 & 74.918 & 67.447 & 63.378 & 66.414 & 38.311 & 37.400 & 59.617 \\ \hline \hline
\end{tabular}}
\caption{Performance with/without LoRA retraining on Vicuna-13B-v1.3.}
\label{lora5}
\end{table*}

\bibliography{aaai2026}